\documentclass[11pt]{article}

\usepackage[preprint]{acl}
\usepackage{booktabs}
\usepackage{times}
\usepackage{latexsym}
\usepackage{amssymb}
\usepackage{amsmath}
\usepackage{enumitem}
\usepackage{tabularx}  
\usepackage{float}
\usepackage{graphicx}
\usepackage{xcolor}
\usepackage{cuted}      
\usepackage{caption} 
\usepackage{algorithm}
\usepackage{algpseudocode}
\usepackage{wasysym}
\newcommand{\mfull}{\CIRCLE}
\newcommand{\mpart}{\LEFTcircle}
\newcommand{\mnone}{\Circle}

\usepackage{needspace}
\usepackage{capt-of}
\usepackage{listings}
\lstdefinestyle{prompt}{basicstyle=\ttfamily\scriptsize,breaklines=true,
breakatwhitespace=false,columns=fullflexible,keepspaces=true,
showstringspaces=false,frame=single,framerule=0.3pt,
xleftmargin=3pt,xrightmargin=3pt,aboveskip=6pt,belowskip=6pt,
postbreak=\mbox{$\hookrightarrow$\space}}

\newcommand{\dn}[1]{{\scriptsize\textcolor{red!70!black}{$\downarrow$#1}}}
\newcommand{\up}[1]{{\scriptsize\textcolor{green!45!black}{$\uparrow$#1}}}
\newcommand{\dns}[1]{{\scriptsize\textcolor{gray}{$\downarrow$#1}}}

\usepackage[T1]{fontenc}

\usepackage[utf8]{inputenc}

\usepackage{microtype}

\usepackage{inconsolata}

\usepackage{graphicx}

\newcommand{\sys}{\textbf{\textsc{TrackMem}}}

\newcommand{\cat}[1]{\textsc{#1}}

\title{TrackMem: Tracking Evolving Beliefs in Long-Horizon Financial Dialogue}

\author{
  Austin M.Y. Cheung\textsuperscript{1,2} \quad
  Harsh Vardhan Gupta\textsuperscript{2} \quad
  Go Nam Lui\textsuperscript{2,3} \\
  Lok Kan Chan\textsuperscript{2} \quad
  Ginny Y. Wong\textsuperscript{4} \quad
  Simon See\textsuperscript{4} \\[4pt]
  \textsuperscript{1}The Hong Kong University of Science and Technology \quad
  \textsuperscript{2}FinCatch Ltd \\
  \textsuperscript{3}City University of Hong Kong \quad
  \textsuperscript{4}NVIDIA AI Technology Center (NVAITC), NVIDIA \\[4pt]
  \texttt{\{austin.cheung, harsh.gupta, harry.lui, lokkan.chan\}@fincatch.io} \\
  \texttt{\{gwong, ssee\}@nvidia.com}
}

\begin{document}
\maketitle
\begin{abstract}

Memory systems for LLM agents are evaluated mostly on social dialogue, where facts accumulate and rarely change, so remembering reduces largely to finding the turn in which something was said. Financial dialogue breaks this assumption: figures are revised, stances shift without being restated, rules stated once govern later questions, and a single company is discussed under several analytical lenses. Memory systems that retrieve the passages most similar to a question, or the most recent ones, therefore return superseded values, miss rules that share no wording with the question, and cannot recover a stance spread across sessions. We present \sys{}, a per-speaker graph memory that tracks the history of figures, separates perspectives by entity and lens, and preserves each speaker's actions in chronological order. To evaluate it, we introduce \textbf{FinLoCoMo}, a benchmark of 15 long-horizon investment conversations with 765 questions in two splits: one following the LoCoMo categories and one targeting four failure modes. With two readers, \sys{} outperforms existing memory systems on both splits. On the failure-mode split it reaches 73.9\% macro accuracy against 69.4\% for the strongest baseline, with the largest gains on questions about revised figures (86.5\% against 73.4\%). \sys{} serves as the memory of a deployed investment assistant. We open-source the implementation at \url{https://github.com/FinCatch/trackmem}.
\end{abstract}

\section{Introduction}\label{sec:intro}
Large Language Model (LLM) agents operating over extended horizons accumulate more information than their context windows can hold, and performance degrades as the context fills \citep{gupta-etal-2024-systematic}. Memory systems address this limitation through persistent storage and retrieval.

Memory systems are evaluated mostly on open-domain social dialogue (LoCoMo, \citealp{maharana-etal-2024-evaluating}; MemBench, \citealp{tan2025membench}), where facts are largely append-only, non-numeric, and remain valid once asserted. Retention thus reduces to locating where a fact was first stated, a task well suited to flat retrieval. In building the memory of a commercial investment assistant, we identified four failure modes specific to financial dialogue.
\begin{figure}[t]
  \centering
  \includegraphics[width=\columnwidth]{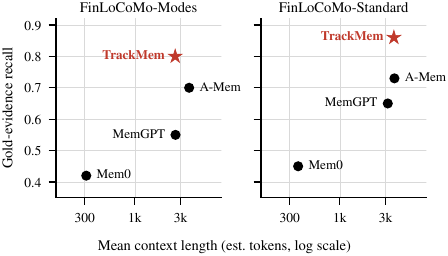}
  \caption{Evidence recall against context length.}
  \label{fig:recall}
\end{figure}
\begin{table}[t]
\centering
\footnotesize
\setlength{\tabcolsep}{3pt}
\begin{tabular}{@{}llcccc@{}}
\toprule
System & Update & {\scriptsize Revision} & {\scriptsize Stance} & {\scriptsize Constraint} & {\scriptsize Aspect} \\
\midrule
Turn RAG & append       & \mnone & \mnone & \mnone & \mnone \\
MemGPT       & self-edit    & \mpart & \mnone & \mpart & \mnone \\
Mem0         & overwrite    & \mpart & \mnone & \mnone & \mnone \\
A-Mem        & note links   & \mnone & \mnone & \mnone & \mnone \\
Zep          & invalidation & \mpart & \mnone & \mnone & \mnone \\
O-Mem        & attributes   & \mnone & \mnone & \mpart & \mpart \\
\midrule
\sys{}       & event time   & \mfull & \mfull & \mfull & \mfull \\
\bottomrule
\end{tabular}
\caption{Failure modes each system addresses by design.
\mfull{}~addressed; \mpart{}~partially; \mnone{}~not addressed.}
\label{tab:coverage}
\end{table}
\textbf{Revision}: quantities are superseded rather than accumulated. \textbf{Stance}: a speaker's stance evolves through dated observations and actions across sessions, requiring reconstruction from the trajectory rather than a single passage. \textbf{Constraint}: a rule stated once may govern later queries with no textual overlap, making similarity- and recency-based retrieval insufficient. \textbf{Aspect}: the same entity is discussed under distinct analytical lenses (e.g., valuation, regulatory risk, and position sizing) that must remain separate. Existing systems address these failure modes at most partially, and none addresses stance (Table~\ref{tab:coverage}).

We propose \sys{}, a per-speaker graph memory whose nodes retain event-time histories, so that it represents what is in force at any date rather than what was last said: revised figures supersede earlier values, stances are reconstructed from dated actions, standing rules remain available to relevant queries, and each analytical lens maintains a distinct view. To evaluate these capabilities, we introduce \textbf{FinLoCoMo}, a benchmark of long-horizon investment dialogues that turns each failure mode into a question category of the same name. We write benchmark categories in small caps (\cat{Revision}, \cat{Stance},
\cat{Constraint}, \cat{Aspect}) and the failure modes in lowercase. Across both splits and both readers, \sys{} delivers the strongest and most consistent performance among the tested memory systems, with its largest gains on \cat{Revision}. It also retrieves more annotated evidence than any baseline at a similar context size (Figure \ref{fig:recall}).

\begin{figure*}[!t]
  \centering
  \includegraphics[width=0.91\textwidth]{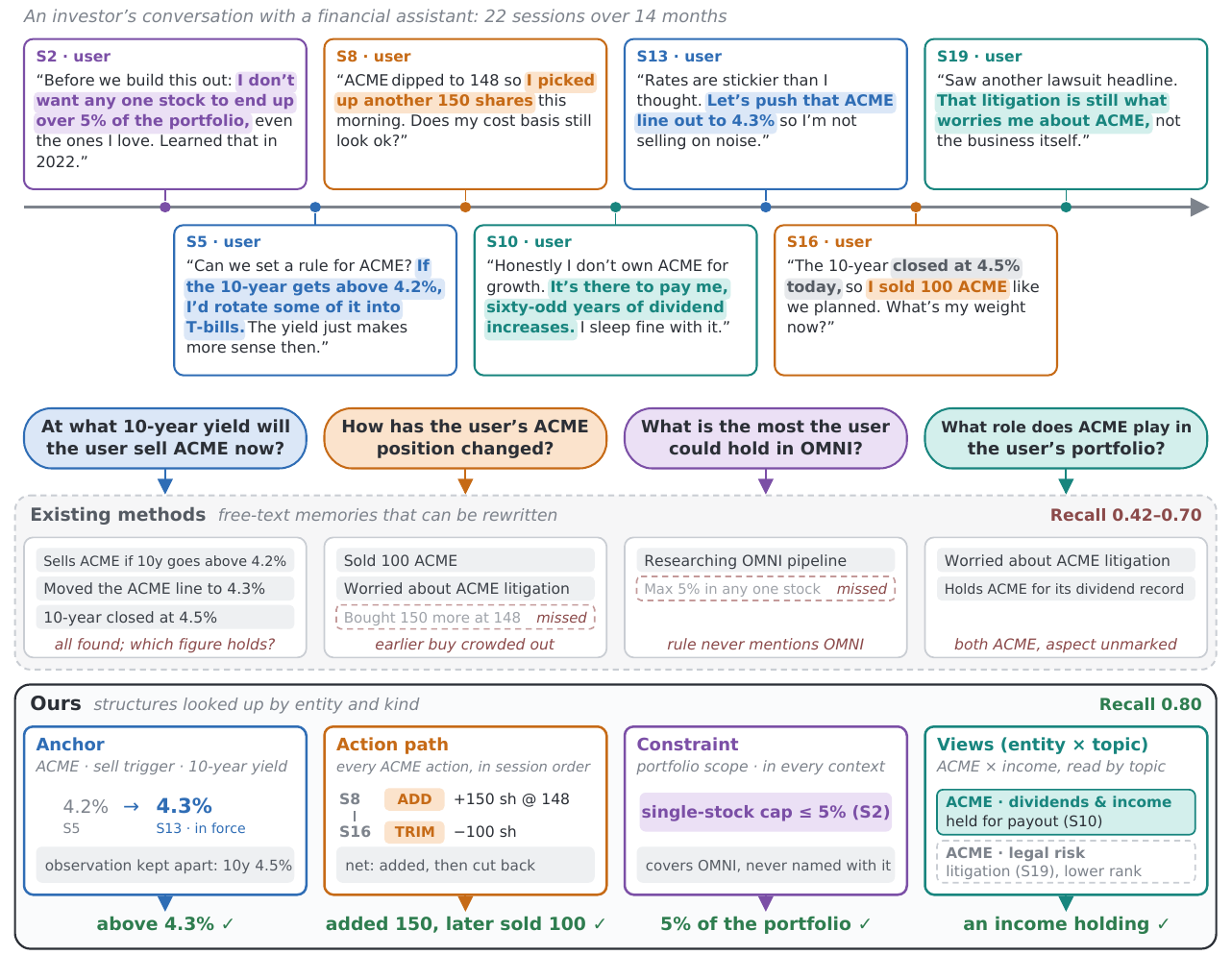}
  \caption{Illustrative example from a long-horizon financial conversation. \sys{} stores each kind of information in its own structure and retrieves it by entity and kind. Recall is evaluated based on FinLoCoMo-Modes.}
  \label{fig:headline}
\end{figure*}

Our contributions are as follows:
\begin{itemize}[leftmargin=*,noitemsep,topsep=2pt]
  \item \textbf{\sys{}}, a per-speaker graph memory that tracks event-time histories of figures, stances, standing rules and lens-specific views, deployed in a commercial investment assistant (\S\ref{sec:deployment}).
  \item \textbf{FinLoCoMo}, a benchmark of 765 questions over 15 long-horizon investment conversations, covering four failure modes and standard LoCoMo categories.
    \item \textbf{An evaluation against five memory systems} under two readers.
    \sys{} is the most accurate on both splits, and outperforms turn-level
    retrieval at a matched retrieval budget and comparable recall.
\end{itemize}

\section{Related Work}
\label{sec:related}
\paragraph{Memory systems.} Memory systems store, organize, and retrieve information from past
interactions to work around the limited context window. They differ
mainly in how memory is represented. MemGPT pages chat logs between the
prompt and external storage, like operating-system memory
\cite{packer2024memgptllmsoperatingsystems}. Mem0 keeps a vector
database of memories, reconciled against new claims through updates or
insertions \cite{chhikara2025mem0}. A-Mem stores linked,
re-annotated atomic notes in the style of a Zettelkasten
\cite{xu2025amemagenticmemoryllm}. Zep uses a bi-temporal knowledge
graph recording when a fact was observed and when it remains valid
\cite{rasmussen2025zep}. O-Mem maintains a user profile with
hierarchical retrieval \cite{wang2025omem}.

None fully addresses any of the four failure modes of \S\ref{sec:intro}
(Table~\ref{tab:coverage}). For revision, Mem0 and O-Mem order facts by when
they were discussed rather than when the event occurred, so recency-based
supersession can keep the wrong value. Zep tracks validity but invalidates a fact  only when a model judges two fact strings to conflict,
so a revised figure can persist beside the old one. For stance and
constraint, all of these systems retrieve atomic items by similarity or by
graph and profile structure, so a stance spread over a trajectory of
statements, or a rule that shares no content with the query, is not
reached. For aspect, only O-Mem indexes memory by topic, which separates topics rather than lenses on the same entity.

\paragraph{Memory benchmarks.}
Existing evaluations rarely exercise these failure modes. Benchmarks with
knowledge updates \citep{wu2025longmemeval,hu2025memoryagentbench} ask for a
fact after it changes, but do not surround the revised value with plans,
hypotheticals and the other speaker's figures, as FinLoCoMo does
(Appendix~\ref{app:benchmarks}).

\section{FinLoCoMo: A Long-Horizon Financial Dialogue Benchmark}
FinLoCoMo extends LoCoMo's generation pipeline
\cite{maharana-etal-2024-evaluating} to investment dialogue. It has 15
two-speaker conversations of 24--28 dated sessions spanning 191--612
days (184--293 in LoCoMo), with two question splits over the same
transcripts (Table~\ref{tab:finlocomo-stats}).

\subsection{Conversation generation}
We add a finance layer and keep LoCoMo's persona-conditioned agents, event graph, session generation with retrieval of earlier facts, and  running summary. Each speaker is seeded with a financial persona of 10--12 first-person lines (role, holding period, leverage and concentration rules, largest holdings). From it, the model writes 80 dated events over 600 days, each linked to the earlier event that caused it; unlinked and off-domain events are discarded. The event graph is the ground truth for what was in force on any date. Each session consumes the agents' next unused events, and stays on financial topics with concrete figures, introducing no new storylines or people. All generation uses MiniMax-M3 (Appendix~\ref{app:persona-and-event-graph}, ~\ref{app:session-gen}).

\subsection{Questions}
The \textbf{Standard} split follows LoCoMo, with 391 \cat{Single-hop}, \cat{Multi-hop}, \cat{Temporal} and \cat{Open-domain} questions and 73 unanswerable \cat{Adversarial} questions. The \textbf{Modes} split turns the four failure modes of \S\ref{sec:intro} into categories designed to mislead similarity- or recency-based retrieval: \cat{Revision} asks for the current value of a revised quantity; \cat{Stance} links a stance change to its trigger and action across sessions; \cat{Constraint} applies a previously stated rule to a figure given elsewhere; and \cat{Aspect} asks for a view under a specified analytical lens or reconciles views across lenses. Candidate questions pass four automatic checks (Appendix~\ref{app:question-gen}); Table~\ref{tab:modes-examples} gives examples.

\begin{table}[!tp]
\centering
\footnotesize
\setlength{\tabcolsep}{5pt}
\begin{tabular}{@{}lrlr@{}}
\toprule
\multicolumn{2}{@{}l}{\textit{Standard split} (464)} &
\multicolumn{2}{l}{\textit{Modes split} (301)} \\
\cmidrule(r){1-2}\cmidrule(l){3-4}
\cat{Single-hop}  & 168 & \cat{Revision}   & 64  \\
\cat{Multi-hop}   & 76  & \cat{Stance}     & 50  \\
\cat{Temporal}    & 74  & \cat{Constraint} & 148 \\
\cat{Open-domain} & 73  & \cat{Aspect}     & 39  \\
\cat{Adversarial} & 73  &            &     \\
\bottomrule
\end{tabular}
\caption{FinLoCoMo questions (765 in total). Conversations average 27.3
sessions of 9.0 turns and 7.9k tokens.}
\label{tab:finlocomo-stats}
\end{table}

\section{\sys{}}
\begin{figure}[t]
  \centering
  \includegraphics[width=0.47\textwidth]{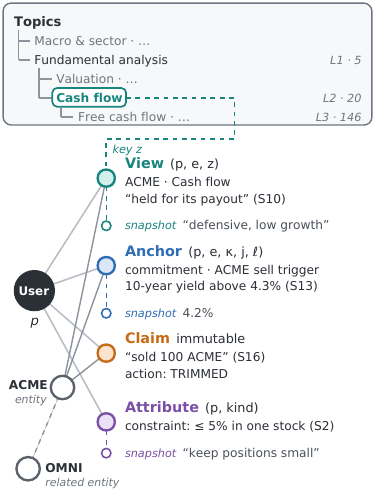}
  \caption{Design of \sys{}}
  \label{fig:trackmem-architecture}
  \label{fig:graph}
\end{figure}

% ---------------------------------------------------------------------------
\subsection{Problem setting}
\label{sec:formulation}
A dialogue between two speakers consists of an ordered sequence of $N$ sessions $S_1,\dots,S_N$. Following \citet{maharana-etal-2024-evaluating}, all turns of session $S_i$ share one timestamp $\tau_i$, with $\tau_1<\dots<\tau_N$.
For each speaker $p$, the
system maintains a memory $\mathcal{G}_p$, updated after every session in
order. Given a question $q$ about $p$, it returns a context
$c=f(q,\mathcal{G}_p)$ from which a fixed reader answers, so the systems
we compare differ only in $\mathcal{G}_p$ and $f$.

% ---------------------------------------------------------------------------
\subsection{Memory representation}
\label{sec:representation}

$\mathcal{G}_p$ is a property graph over the entities $p$ discusses,
organised by a fixed taxonomy of financial topics
(Appendix~\ref{app:taxonomy}), with four node types
(Figure~\ref{fig:graph}).
 
\textbf{Views} hold $p$'s current understanding of an entity $e$ under one
analytical lens, as a short summary with an optional sentiment. A view is
keyed by $(p,e,z)$ with $z$ a topic, so the valuation, regulatory risk and
sizing of one security have separate histories.
 
\textbf{Anchors} hold figures as typed values. A \emph{commitment} is a
decision $p$ may revise (a price target, a position cap); an
\emph{observation} is true when stated (a market level, an executed fill).
An anchor is keyed by $(p,e,\kappa,j,\ell)$: its kind $\kappa$, a slot $j$
that separates different quantities of the same kind
(\S\ref{sec:construction}), and a period $\ell$. For a commitment the
period is empty, so a later value replaces the earlier one; for an
observation it is the date stated, so readings accumulate.
 
\textbf{Attributes} hold standing properties of $p$, such as risk tolerance
or a portfolio-wide rule (``no leverage''), keyed by $(p,\kappa)$ with kind $\kappa$.

\textbf{Claims} are single assertions from one turn and are never
revised; a claim recording something $p$ did carries an action kind
(\texttt{added}, \texttt{trimmed}, \dots).
 
Views, anchors and attributes keep every earlier state in a history $H(n)$,
each with its validity interval in \emph{event} time, when the
conversation took place rather than when the node was written. Fields and
vocabularies are listed in Appendix~\ref{app:rep-and-cons}.
 
\subsection{Memory construction}
\label{sec:construction}
Each session updates $\mathcal{G}_p$ once (Algorithm~\ref{alg:write}).
A schema-constrained extraction maps the session onto closed
vocabularies and links its entities to those already in $\mathcal{G}_p$.
The kind of a figure does not identify the quantity it measures. We
compare each new commitment $a$ with the stored anchors
$Q(a)$ that share its entity, kind, unit and period, and a resolver
$\psi$ selects the one that $a$ revises, if any. When the match is
ambiguous, $\psi$ assigns a new slot, because merging two distinct
quantities would overwrite a valid value, while keeping them apart
costs only a duplicate. Each node is then written under its key, and a
changed state closes the previous one at $\tau_k$. The current value of
each quantity is therefore determined by its key rather than by a model
judgement (Appendix~\ref{app:rep-and-cons}).

\begin{algorithm}[t]
\footnotesize
\caption{Writing session $S_k$ into $\mathcal{G}_p$.}
\label{alg:write}
\begin{algorithmic}[1]
\State $(Z, A, C, P) \gets \textsc{Extract}(S_k)$
\State map the entities in $Z, A, C$ onto those in $\mathcal{G}_p$
\For{each commitment $a \in A$}
  \State $j(a) \gets \psi\big(a, Q(a)\big)$ \Comment{revised slot, or new}
\EndFor
\For{each view $(e,z) \in Z$ and attribute in $P$}
  \State new text $\gets \textsc{Synthesise}(\text{stored text}, S_k)$
\EndFor
\For{each view, anchor and attribute $n$ not yet updated by $S_k$}
  \If{$n$ is new or its state changed}
    \State append $\big(\mathrm{state}(n), [t^{\mathrm{from}}(n), \tau_k)\big)$ to $H(n)$
    \State $\mathrm{state}(n) \gets$ new state;\ \ $t^{\mathrm{from}}(n) \gets \tau_k$
  \EndIf
\EndFor
\State add the claims $C$ \Comment{immutable}
\end{algorithmic}
\end{algorithm}

% ---------------------------------------------------------------------------
\subsection{Retrieval}
\label{sec:retrieval}
A single model call maps the question to the entities $E_q$ and topics
$Z_q$ it mentions and to a question type. Each node type is then retrieved through its own
channel. Current views, past views and claims are ranked by similarity
to the question. Current views additionally receive fixed bonuses for a
matching entity and topic, must exceed a similarity threshold $\theta$,
and are limited per entity. Anchors with their value histories, the
trajectory of $p$'s claims about $E_q$ in event order, and all
attributes are retrieved by exact lookup, independent of the question's
wording (Appendix~\ref{app:config}). The retrieved items form the
context $c=[U;A;V;\bar{V};P;T]$: verbatim turns $U$, anchors $A$, current
views $V$, past views $\bar{V}$, the profile $P$ and the trajectory $T$.
$U$ holds the source turns of the retrieved claims and trajectory, plus
the source turns of other retrieved items most similar to the question.

\section{Experiments}
\label{sec:experiments}

\subsection{Setup}
\label{sec:setup}
We compare \sys{} with five memory systems: MemGPT
\citep{packer2024memgptllmsoperatingsystems}, Mem0 \citep{chhikara2025mem0},
A-Mem \citep{xu2025amemagenticmemoryllm}, Zep, via its open-source engine
Graphiti \citep{rasmussen2025zep}, and O-Mem \citep{wang2025omem}, each run
from its released code with default settings. Each system builds its memory
once from the 15 FinLoCoMo conversations with \texttt{gemini-3.1-flash-lite},
embedding with \texttt{gemini-embedding-2}, then retrieves context once per
question. The contexts are answered under the same prompt by two readers at
temperature 0: \texttt{gemini-3.1-flash-lite}
and \texttt{nemotron-3.5-lightning} with reasoning, so recall and context
length do not depend on the reader. An LLM judge (MiniMax-M3) grades each
answer as correct (1), partial (0.5) or incorrect (0). Tokens are estimated as characters / 4. \emph{Macro} weights the four categories of a split equally; \emph{Micro} weights each answerable question equally.

\subsection{Main results}
\label{sec:main-results}

\begin{table*}[!t]
\centering
\footnotesize
\setlength{\tabcolsep}{2.7pt}
\renewcommand{\arraystretch}{1.06}
\newcommand{\hc}[1]{{\scriptsize\cat{#1}}}
\begin{tabular}{@{}l cc cccccc cccccc@{}}
\toprule
\multicolumn{15}{@{}l}{\textbf{FinLoCoMo-Modes}} \\
\midrule
 & \multicolumn{2}{c}{Retrieval} & \multicolumn{6}{c}{Gemini 3.1 Flash Lite} & \multicolumn{6}{c}{Nemotron 3.5 Lightning} \\
\cmidrule(lr){2-3}\cmidrule(lr){4-9}\cmidrule(l){10-15}
Method & Recall & Tokens & \hc{Rev.} & \hc{Stance} & \hc{Con.} & \hc{Aspect} & Macro & Micro & \hc{Rev.} & \hc{Stance} & \hc{Con.} & \hc{Aspect} & Macro & Micro \\
 & & & {\scriptsize(64)} & {\scriptsize(50)} & {\scriptsize(148)} & {\scriptsize(39)} & & {\scriptsize(301)} & {\scriptsize(64)} & {\scriptsize(50)} & {\scriptsize(148)} & {\scriptsize(39)} & & {\scriptsize(301)} \\
\midrule
Mem0            & 0.42 & 311   & 54.0 & 12.8 & 66.3 & 27.0 & 40.0 & 50.0 & 48.4 & 15.0 & 47.3 & 32.1 & 35.7 & 40.2 \\
A-Mem           & 0.70 & 3,687 & 69.4 & 56.1 & \textbf{79.3} & 59.2 & 66.0 & 70.7 & 76.6 & 61.0 & \textbf{80.7} & 65.4 & 70.9 & 74.6 \\
MemGPT          & 0.55 & 2,656 & 73.4 & 55.1 & 76.5 & \textbf{72.4} & 69.4 & 71.8 & 80.5 & 66.0 & 76.4 & \textbf{83.3} & 76.5 & 76.4 \\
Graphiti (Zep)  & 0.48 & 318   & 40.6 & 14.0 & 61.1 & 26.9 & 35.7 & 44.5 & 42.2 & 11.0 & 41.6 & 21.8 & 29.1 & 34.1 \\
O-Mem           & 0.33 & 943   & 60.2 & 49.0 & 65.2 & 35.9 & 52.6 & 57.6 & 54.7 & 51.0 & 49.0 & 42.3 & 49.2 & 49.7 \\
\sys{}          & \textbf{0.80} & 2,636 & \textbf{86.5} & \textbf{61.5} & 79.1 & 68.4 & \textbf{73.9} & \textbf{76.4} & \textbf{84.4} & \textbf{71.0} & 77.4 & 74.4 & \textbf{76.8} & \textbf{77.4} \\
\midrule
\emph{Full context} & 1.00 & 10,429 & 82.0 & 63.0 & 76.0 & 63.5 & 71.1 & 73.6 & 82.8 & 79.0 & 81.8 & 70.5 & 78.5 & 80.1 \\
\emph{Gold evidence}   & 1.00 & --     & 88.3 & 56.0 & 78.0 & 75.6 & 74.5 & 76.3 & 80.5 & 66.0 & 64.5 & 71.8 & 70.7 & 69.1 \\
\midrule[\heavyrulewidth]
\multicolumn{15}{@{}l}{\textbf{FinLoCoMo-Standard}} \\
\midrule
 & \multicolumn{2}{c}{Retrieval} & \multicolumn{6}{c}{Gemini 3.1 Flash Lite} & \multicolumn{6}{c}{Nemotron 3.5 Lightning} \\
\cmidrule(lr){2-3}\cmidrule(lr){4-9}\cmidrule(l){10-15}
Method & Recall & Tokens & \hc{Single} & \hc{Multi} & \hc{Temp.} & \hc{Open} & Macro & Micro & \hc{Single} & \hc{Multi} & \hc{Temp.} & \hc{Open} & Macro & Micro \\
 & & & {\scriptsize(168)} & {\scriptsize(76)} & {\scriptsize(74)} & {\scriptsize(73)} & & {\scriptsize(391)} & {\scriptsize(168)} & {\scriptsize(76)} & {\scriptsize(74)} & {\scriptsize(73)} & & {\scriptsize(391)} \\
\midrule
Mem0            & 0.45 & 367   & 53.7 & 44.4 & 25.7 & 42.4 & 41.6 & 44.6 & 54.5 & 39.5 & 15.5 & 42.5 & 38.0 & 41.9 \\
A-Mem           & 0.73 & 3,714 & 84.5 & 65.5 & 40.3 & \textbf{52.2} & 60.6 & 66.9 & 86.6 & 69.7 & 41.2 & 55.5 & 63.3 & 68.9 \\
MemGPT          & 0.65 & 3,165 & \textbf{88.6} & 72.3 & 48.6 & 47.8 & 64.3 & 70.5 & 90.2 & \textbf{76.3} & 45.3 & 54.1 & 66.5 & 72.3 \\
Graphiti (Zep)  & 0.55 & 315   & 42.9 & 41.4 & 35.1 & 37.0 & 39.1 & 40.0 & 36.0 & 38.2 & 25.7 & 26.7 & 31.6 & 32.7 \\
O-Mem           & 0.44 & 1,099 & 64.6 & 59.9 & 25.7 & 48.6 & 49.7 & 53.3 & 64.9 & 46.1 & 24.3 & 46.6 & 45.5 & 50.1 \\
\sys{}          & \textbf{0.86} & 3,666 & 85.5 & \textbf{75.0} & \textbf{54.9} & 51.4 & \textbf{66.7} & \textbf{71.4} & \textbf{90.5} & 71.7 & \textbf{56.1} & \textbf{57.5} & \textbf{69.0} & \textbf{74.2} \\
\midrule
\emph{Full context} & 1.00 & 10,512 & 83.7 & 67.3 & 43.2 & 47.7 & 60.5 & 66.5 & 87.2 & 75.7 & 58.8 & 63.0 & 71.2 & 75.1 \\
\emph{Gold evidence}   & 1.00 & --     & 94.9 & 80.9 & 64.2 & 50.7 & 72.7 & 78.1 & 86.9 & 80.3 & 58.8 & 68.5 & 73.6 & 76.9 \\
\bottomrule
\end{tabular}
\caption{Main results (\%). \cat{Rev.}\,=\,\cat{Revision}, \cat{Con.}\,=\,\cat{Constraint}.
\emph{Recall}: share of annotated evidence turns in the context. \emph{Micro} is over successfully judged questions (range from 0 -- 18 unjudged per cell).}
\label{tab:main}
\end{table*}

\paragraph{\sys{} is the most accurate memory system with both readers.}
It leads MemGPT, the strongest baseline, by 4.5 and 0.3 macro on Modes and
by 2.4 and 2.5 on Standard (Table~\ref{tab:main}). \sys{} also shows the reader the most gold evidence (recall 0.80 and 0.86, against at most 0.70 and 0.73), with a
quarter to a third of the full context's tokens.

\paragraph{The gains are on the failure modes.}
With both readers \sys{} is the best memory system on \cat{Revision} and
\cat{Stance}; MemGPT leads on \cat{Aspect}, and A-Mem on \cat{Constraint},
marginally with Gemini. On Standard, \sys{} leads on \cat{Temporal}, which
requires ordering events across sessions. Mem0, Zep and O-Mem return at
most 1.1k tokens with recall of at most 0.55, and trail \sys{} on every
category.

\paragraph{The reader decides how much full context recovers.}
With Gemini, full context trails \sys{} on both splits despite containing
every gold turn: most quantities are stated several times, as plans,
revisions and the other speaker's figures, and the reader often picks a
superseded one (Appendix~\ref{app:examples}). The reasoning reader recovers
more from the whole transcript and leads overall, most clearly on
\cat{stance}, whose evidence is spread over many turns. Gold evidence is not
a ceiling either: with Nemotron it scores below \sys{} on Modes and
declines 60 of the 301 questions, as annotated turns lack surrounding dates and actions.

\paragraph{\sys{} abstains with high precision.}
Treating a decline as a prediction that the question is unanswerable,
\sys{} has the highest precision with both readers (91.5
and 87.8; Appendix~\ref{app:results}) but declines fewer unanswerable questions, so its F1 matches MemGPT's with Gemini and trails it with Nemotron.

% ---------------------------------------------------------------------------
\subsection{Where the advantage comes from}
\label{sec:analysis}

\paragraph{Structure beats retrieval at an equal budget.}
A turn-level RAG baseline that embeds every turn and returns the 27 most
similar to the question, the number of turns \sys{} shows, reaches nearly
the same recall (0.82 against 0.80) but scores 5.5 points lower in macro
accuracy and 16.3 lower on \cat{Revision} (Appendix~\ref{app:results}). The
evidence turns are in its context, but nothing tells the reader which
stated value is current.

\paragraph{Upstream ablation.}
We rebuild the memory without one extracted component and rerun
retrieval (Table \ref{tab:ablation}). Removing claims costs the most although claims are never shown to the reader, and the share of gold turns retrieved falls from 79.0\% to 65.5\%, so claims serve to find evidence. 

\paragraph{Context ablation.}
We remove either the verbatim turns or the anchors from the rendered context after retrieval, keeping retrieval fixed. Without verbatim turns, accuracy falls by 13.6
points, and without anchors by 3.3. 

\begin{table}[t]
\centering
\footnotesize
\setlength{\tabcolsep}{10pt}
\begin{tabular}{@{}lcc@{}}
\toprule
 & Modes & Standard \\
\midrule
\multicolumn{3}{@{}l}{\textit{Upstream ablation}} \\
Full memory   & 74.4             & 70.3 \\
\quad $-C$    & 70.6\,\dn{3.8}   & 68.8\,\dns{1.5} \\
\quad $-A$    & 73.2\,\dns{1.2}  & 69.3\,\dns{1.0} \\
\quad $-P$    & 74.0\,\dns{0.4}        & 68.9\,\dns{1.4} \\
\midrule
\multicolumn{3}{@{}l}{\textit{Context ablation}} \\
All sections  & 75.9             & -- \\
\quad $-U$    & 62.3\,\dn{13.6}  & -- \\
\quad $-A$    & 72.6\,\dn{3.3}   & -- \\
\bottomrule
\end{tabular}
\caption{Micro accuracy after removing one component. Each
block is a separate build and judge run, so its first row differs from Table \ref{tab:main}.}
\label{tab:ablation}
\end{table}

\section{Deployment}
\label{sec:deployment}

\paragraph{Architecture and reliability.}
\sys{} runs as four services connected by Redis streams: an API that
accepts turns and answers queries, a buffer that groups turns per user,
a dispatcher that makes all write-path model calls, and a writer that
updates the Neo4j graph (Appendix~\ref{app:deployment};
Table~\ref{tab:production}). Each stage acknowledges a message only
after storing its result, and writes are idempotent, so at-least-once
delivery is safe; batches that repeatedly fail validation are held in a dead-letter queue for replay. Queries are restricted to the user's subgraph, and the
recommendation layer reads only per-user engagement counts on shared
entity and topic nodes.

\paragraph{Cost and ordering.}
A user’s turns are buffered into batches of up to 2,048 tokens or 20 messages, and a six-hour inactivity timeout. Extraction requests are submitted through the provider’s asynchronous batch interface. Users are hashed to shards, and one worker writes each shard’s output stream sequentially. At query time, the router is the only generative model call, alongside embedding and graph operations; channel-specific keep the context at about 2.6k tokens on FinLoCoMo-Modes.

\paragraph{Lessons from deployment.}
Four production failures shaped the implementation.
\emph{Attribution.} Figures in the assistant's replies were recorded as
the user's commitments; a figure now enters the user's memory only if
the user states or adopts it.
\emph{Scale.} Users wrote figures in different scales or format (0.08 and 8\%); values are now normalised to one scale per unit.
\emph{Scope.} Portfolio-wide rules were stored on each company
discussed, so a later change reached only some of them; they are now
stored once, as a user attribute.
\emph{Non-subjects.} Uploaded files, drafts and the assistant itself
were extracted as investment entities; extraction now admits only
subjects on which the user can hold a view.

\paragraph{Evaluated and deployed systems.}
The evaluation uses the production write and read paths, with sessions
in place of batches. Production also links entities to a proprietary
market-event graph and learns the view-score weights $\lambda$; the hand-set weights of Appendix~\ref{app:config} serve as its cold start and are used in all reported results.

\begin{table}[t]
\centering
\footnotesize
\begin{tabular}{@{}lr@{}}
\toprule
In production since                  & July 2026 \\
Ingested turns               & 13,781 \\
Nodes per user  & 604\\
Estimated model write cost per batch & $\approx\$0.005$ \\
Ingestion delay (median)             & 19.0\,min \\
Estimated query latency              & 3--5\,s \\
Batches entering dead-letter queue   & 0.13\% \\
\bottomrule
\end{tabular}
\caption{\sys{} deployment (Oct 2026).}
\label{tab:production}
\end{table}

\newpage
\section{Limitations}
\sys{} relies on a language model to build its memory. Extraction, quantity
resolution and view synthesis are model judgements, so an error made when a
session is written, such as a missed figure, a wrong entity or two quantities
merged into one slot, persists until a later session revises it. Writing
memory also costs model calls at ingestion, which retrieval over raw turns
avoids. The graph stores turn identifiers rather than turn text, so the
verbatim section of the context requires access to the original transcript.

Our evaluation covers one domain and one language, with two speakers per
conversation, and conversations short enough to fit in the reader's context.
Behaviour over longer histories, other domains and many-party dialogue is
untested. Accuracy is graded by an LLM judge, whose verdicts vary between
runs by about two macro points, and the configuration of \sys{} was selected
on the evaluation questions, while the baselines were run with their released
settings.

\clearpage
\newpage
\bibliography{custom}

@misc{packer2024memgptllmsoperatingsystems,
      title={MemGPT: Towards LLMs as Operating Systems}, 
      author={Charles Packer and Sarah Wooders and Kevin Lin and Vivian Fang and Shishir G. Patil and Ion Stoica and Joseph E. Gonzalez},
      year={2024},
      eprint={2310.08560},
      archivePrefix={arXiv},
      primaryClass={cs.AI},
      url={https://arxiv.org/abs/2310.08560}, 
}

@misc{xu2025amemagenticmemoryllm,
      title={A-MEM: Agentic Memory for LLM Agents}, 
      author={Wujiang Xu and Zujie Liang and Kai Mei and Hang Gao and Juntao Tan and Yongfeng Zhang},
      year={2025},
      eprint={2502.12110},
      archivePrefix={arXiv},
      primaryClass={cs.CL},
      url={https://arxiv.org/abs/2502.12110}, 
}

@article{Huang_2026,
   title={A Survey on Retrieval-Augmented Text Generation for Large Language Models},
   volume={58},
   ISSN={1557-7341},
   url={http://dx.doi.org/10.1145/3805774},
   DOI={10.1145/3805774},
   number={12},
   journal={ACM Computing Surveys},
   publisher={Association for Computing Machinery (ACM)},
   author={Huang, Yizheng and Huang, Jimmy Xiangji},
   year={2026},
   month=May, pages={1–38} }

@misc{singh2026agenticretrievalaugmentedgenerationsurvey,
      title={Agentic Retrieval-Augmented Generation: A Survey on Agentic RAG}, 
      author={Aditi Singh and Abul Ehtesham and Saket Kumar and Tala Talaei Khoei and Athanasios V. Vasilakos},
      year={2026},
      eprint={2501.09136},
      archivePrefix={arXiv},
      primaryClass={cs.AI},
      url={https://arxiv.org/abs/2501.09136}, 
}

@misc{edge2025localglobalgraphrag,
      title={From Local to Global: A Graph RAG Approach to Query-Focused Summarization}, 
      author={Darren Edge and Ha Trinh and Newman Cheng and Joshua Bradley and Alex Chao and Apurva Mody and Steven Truitt and Dasha Metropolitansky and Robert Osazuwa Ness and Jonathan Larson},
      year={2025},
      eprint={2404.16130},
      archivePrefix={arXiv},
      primaryClass={cs.CL},
      url={https://arxiv.org/abs/2404.16130}, 
}

@misc{gao2024retrievalaugmentedgenerationlargelanguage,
      title={Retrieval-Augmented Generation for Large Language Models: A Survey}, 
      author={Yunfan Gao and Yun Xiong and Xinyu Gao and Kangxiang Jia and Jinliu Pan and Yuxi Bi and Yi Dai and Jiawei Sun and Meng Wang and Haofen Wang},
      year={2024},
      eprint={2312.10997},
      archivePrefix={arXiv},
      primaryClass={cs.CL},
      url={https://arxiv.org/abs/2312.10997}, 
}

@article{chhikara2025mem0,
  title={Mem0: Building production-ready ai agents with scalable long-term memory},
  author={Chhikara, Prateek and Khant, Dev and Aryan, Saket and Singh, Taranjeet and Yadav, Deshraj},
  journal={arXiv preprint arXiv:2504.19413},
  year={2025}
}

@article{rasmussen2025zep,
  title={Zep: a temporal knowledge graph architecture for agent memory},
  author={Rasmussen, Preston and Paliychuk, Pavlo and Beauvais, Travis and Ryan, Jack and Chalef, Daniel},
  journal={arXiv preprint arXiv:2501.13956},
  year={2025}
}

@inproceedings{maharana-etal-2024-evaluating,
    title = "Evaluating Very Long-Term Conversational Memory of {LLM} Agents",
    author = "Maharana, Adyasha  and
      Lee, Dong-Ho  and
      Tulyakov, Sergey  and
      Bansal, Mohit  and
      Barbieri, Francesco  and
      Fang, Yuwei",
    editor = "Ku, Lun-Wei  and
      Martins, Andre  and
      Srikumar, Vivek",
    booktitle = "Proceedings of the 62nd Annual Meeting of the Association for Computational Linguistics (Volume 1: Long Papers)",
    month = aug,
    year = "2024",
    address = "Bangkok, Thailand",
    publisher = "Association for Computational Linguistics",
    url = "https://aclanthology.org/2024.acl-long.747/",
    doi = "10.18653/v1/2024.acl-long.747",
    pages = "13851--13870",
}

@inproceedings{tan2025membench,
  title={Membench: Towards more comprehensive evaluation on the memory of llm-based agents},
  author={Tan, Haoran and Zhang, Zeyu and Ma, Chen and Chen, Xu and Dai, Quanyu and Dong, Zhenhua},
  booktitle={Findings of the Association for Computational Linguistics: ACL 2025},
  pages={19336--19352},
  year={2025}
}

@article{wang2025omem,
  title     = {O-Mem: Omni Memory System for Personalized, Long Horizon, Self-Evolving Agents},
  author    = {Wang, Piaohong and Tian, Motong and Li, Jiaxian and Liang, Yuan and Wang, Yuqing and Chen, Qianben and Wang, Tiannan and Lu, Zhicong and Ma, Jiawei and Jiang, Yuchen Eleanor and Zhou, Wangchunshu},
  journal   = {arXiv preprint arXiv:2511.13593},
  year      = {2025},
  doi       = {10.48550/arXiv.2511.13593},
  url       = {https://arxiv.org/abs/2511.13593}
}

@inproceedings{gupta-etal-2024-systematic,
    title = "Systematic Evaluation of Long-Context {LLM}s on Financial Concepts",
    author = "Gupta, Lavanya  and
      Sharma, Saket  and
      Zhao, Yiyun",
    editor = "Dernoncourt, Franck  and
      Preo{\c{t}}iuc-Pietro, Daniel  and
      Shimorina, Anastasia",
    booktitle = "Proceedings of the 2024 Conference on Empirical Methods in Natural Language Processing: Industry Track",
    month = nov,
    year = "2024",
    address = "Miami, Florida, US",
    publisher = "Association for Computational Linguistics",
    url = "https://aclanthology.org/2024.emnlp-industry.88/",
    doi = "10.18653/v1/2024.emnlp-industry.88",
    pages = "1163--1175"
}

@inproceedings{wu2025longmemeval,
  title     = {{LongMemEval}: Benchmarking Chat Assistants on Long-Term Interactive Memory},
  author    = {Wu, Di and Wang, Hongwei and Yu, Wenhao and Zhang, Yuwei and Chang, Kai-Wei and Yu, Dong},
  booktitle = {The Thirteenth International Conference on Learning Representations},
  year      = {2025}
}

@article{hu2025memoryagentbench,
  title   = {Evaluating Memory in {LLM} Agents via Incremental Multi-Turn Interactions},
  author  = {Hu, Yuanzhe and Wang, Yu and McAuley, Julian},
  journal = {arXiv preprint arXiv:2507.05257},
  year    = {2025}
}

@inproceedings{chen-etal-2021-finqa,
  title     = {{FinQA}: A Dataset of Numerical Reasoning over Financial Data},
  author    = {Chen, Zhiyu and Chen, Wenhu and Smiley, Charese and Shah, Sameena and Borova, Iana and Langdon, Dylan and Moussa, Reema and Beane, Matt and Huang, Ting-Hao and Routledge, Bryan and Wang, William Yang},
  booktitle = {Proceedings of the 2021 Conference on Empirical Methods in Natural Language Processing},
  pages     = {3697--3711},
  year      = {2021},
  url       = {https://aclanthology.org/2021.emnlp-main.300}
}

@inproceedings{chen-etal-2022-convfinqa,
  title     = {{ConvFinQA}: Exploring the Chain of Numerical Reasoning in Conversational Finance Question Answering},
  author    = {Chen, Zhiyu and Li, Shiyang and Smiley, Charese and Ma, Zhiqiang and Shah, Sameena and Wang, William Yang},
  booktitle = {Proceedings of the 2022 Conference on Empirical Methods in Natural Language Processing},
  pages     = {6279--6292},
  year      = {2022},
  url       = {https://aclanthology.org/2022.emnlp-main.421}
}

@article{islam2023financebench,
  title   = {{FinanceBench}: A New Benchmark for Financial Question Answering},
  author  = {Islam, Pranab and Kannappan, Anand and Kiela, Douwe and Qian, Rebecca and Scherrer, Nino and Vidgen, Bertie},
  journal = {arXiv preprint arXiv:2311.11944},
  year    = {2023}
}

\newpage

\clearpage
\appendix
\section*{Appendix}
\section{Extended related works}
\subsection{Retrieval-Augmented Generation}\label{app:rag}
Retrieval-Augmented Generation (RAG) augments LLMs with externally retrieved
evidence. Naïve RAG chunks source documents, embeds them in a vector store,
and inserts the top-$k$ semantically similar chunks into the prompt
\citep{gao2024retrievalaugmentedgenerationlargelanguage}. Advanced RAG
improves retrieval quality through query reformulation, expansion, and index
refinement before search, as well as re-ranking and filtering afterward
\citep{Huang_2026}. Agentic RAG further delegates retrieval decisions to an
LLM, allowing it to determine when and what to retrieve during reasoning
\citep{singh2026agenticretrievalaugmentedgenerationsurvey}. GraphRAG targets
global questions by constructing an entity knowledge graph, summarizing its
hierarchical communities, and combining partial answers generated from these
summaries \citep{edge2025localglobalgraphrag}. Across these variants,
however, retrieval remains largely static: they do not maintain or revise an
evolving belief state as new information arrives over time.

\subsection{Long-Context LLM Evaluation on Financial Concepts}\label{app:long-context-finance}
\citet{gupta-etal-2024-systematic} evaluate GPT-4-family models on financial-news retrieval tasks of increasing context length and complexity. They find that a model’s advertised context window does not imply reliable use of the full context: GPT-4o’s F1 on a simple company-identification task falls from 0.99 at 4K tokens to 0.40 at 128K tokens.

Degradation is more severe for multi-concept tasks, such as jointly identifying a company and its sentiment, where performance approaches zero beyond roughly 32K tokens. The authors also report instruction-following and formatting failures at long context lengths, as well as sensitivity to prompt placement and minor formatting changes. These results indicate that long-context LLMs may require retrieval, controlled context construction, and explicit abstention mechanisms in financial applications.

\subsection{Memory and Financial QA Benchmarks}\label{app:benchmarks}
Closest to FinLoCoMo are benchmarks with knowledge updates: LongMemEval and
MemoryAgentBench include questions whose answer changes during the
interaction \citep{wu2025longmemeval,hu2025memoryagentbench}. FinLoCoMo
differs in what surrounds the update. A revised figure co-occurs with plans,
hypotheticals and the other speaker's figures for the same quantity, so the
latest mention is often not the answer (Table~\ref{tab:modes-examples}), and
the benchmark also targets stance trajectories and analytical lenses.
Financial QA benchmarks evaluate numerical reasoning over a single filing or
conversation rather than memory across sessions
\citep{chen-etal-2021-finqa,chen-etal-2022-convfinqa,islam2023financebench}.

\section{FinLoCoMo Construction}
\subsection{Personas and event graph}\label{app:persona-and-event-graph}

\paragraph{Personas.} Each speaker is seeded with 10--12 first-person
attribute lines describing a financial life: role, market focus,
holding period, leverage and concentration rules, risk appetite,
rebalancing habit, and largest holdings (e.g., ``I run a long-only
book with no leverage and strict single-name concentration limits'').
Roles range from professionals (portfolio manager, sell-side bank
analyst, rates trader, currency-overlay manager) to serious private
investors.

\paragraph{Event graphs.} From the persona, the model writes 80 dated
events per speaker over a 600-day window, each linked to the earlier
event that caused it; events with no causal link are discarded. For
example, a quarterly result raises conviction, which leads to a larger
position. The graph therefore records each revision of a quantity held
about an entity, and each chain from observation to change of
conviction to action. It serves as the ground truth for what was in
force at any date. A separate LLM call acts as a classifier that
removes off-domain events before any dialogue is written.
\subsection{Session generation}\label{app:session-gen}
\paragraph{Sessions.} Each session is generated iteratively by
consuming the next unused event from each agent's own timeline. The
speaking agent sees its persona, the date, the running summary, its
own events since the last session, and two facts per speaker retrieved
by embedding similarity from earlier sessions. During generation, the
model is instructed to keep the conversation to financial topics, to
use concrete figures rather than qualitative language, and to avoid
introducing new personal storylines or new named people.

\subsection{Question generation}\label{app:question-gen}
\paragraph{Quality control.}
A candidate question is accepted only if it passes four checks:
(i)~structural checks against the conversation, such as the sessions
spanned, the number of revisions and the speaker named; (ii)~a blind
check, in which MiniMax-M3 answers from the full context, without the
gold, and the same model must judge that answer correct; (iii)~rejection
of the candidate if any of five constant answers is also judged correct;
and (iv)~deduplication against the Standard split
(Appendix~\ref{app:locomo-statistics}).
 
\subsection{Human validation}\label{app:human-val}
We checked a stratified sample of 120 questions (20 per Modes
category and 8 per Standard category, including unanswerable ones) for
whether the gold answer is correct, the cited evidence suffices, and the
question fits its category. 98.3\% of gold answers, 96.7\% of evidence sets
and 94.2\% of category assignments were judged valid. We also checked the MiniMax-M3 judge on 24
answers balanced across verdicts and agreed with 24/24 verdicts.
 
\subsection{Statistics}\label{app:locomo-statistics}
 
\paragraph{Conversations.}
FinLoCoMo has 15 conversations between two speakers each, with 410
sessions in total (Table~\ref{tab:conv-stats}). Events are generated over
a 600-day horizon, and each session is dated one or two days after the
event it consumes, so conversations span 191--612 days. The full context
shown to the reader averages 10,429 tokens on Modes and 10,512 on
Standard; the difference from the 7.9k tokens of turn text is the date and
speaker prefix on each turn. Turn text contains about 307 numeric mentions
per conversation (range 200--419, counting repeats) and about 21 distinct
tickers (range 12--32), counted as uppercase strings of two to five
letters after removing common abbreviations. Each \cat{Revision} question
cites 2.5 distinct values on average, and at most 6; whether its latest
mention is the gold is labelled in the human check
(Appendix~\ref{app:human-val}).
 
\begin{table}[h]
\centering
\small
\begin{tabular}{@{}lrrr@{}}
\toprule
 & Mean & Min & Max \\
\midrule
Sessions           & 27.3 & 24   & 28   \\
Turns              & 247  & 196  & 282  \\
Turns per session  & 9.0  & 2    & 22   \\
Days spanned       & 411  & 191  & 612  \\
Turn text (tokens) & 7.9k & 5.6k & 9.5k \\
\bottomrule
\end{tabular}
\caption{Per-conversation statistics of FinLoCoMo.}
\label{tab:conv-stats}
\end{table}
 
\paragraph{Questions.}
Table~\ref{tab:question-stats} reports, per category, the number of
annotated evidence turns, the number of sessions they come from, and the
session span, the distance from the earliest cited session to the latest
(0 when the evidence lies in one session). Eight of the 168
\cat{Single-hop} questions cite more than one session, as in LoCoMo.
 
Of the 148 \cat{Constraint} questions, 90 place the rule and another cited
figure in different sessions (median gap 4, mean 7.6, maximum 26), and 25
place them in the same session. The structural check, which tests wording
overlap rather than the number of cited turns, does not exclude two
further groups: 22 questions cite only the rule turn, with the figure in
that turn or in the question, and 11 leave the rule turn unlabelled, of
which 9 cite more than one turn. All of these 25 + 33 questions passed the
same checks as the rest of the category: overlap between the question and
the rule stays within the structural cap, MiniMax-M3 reproduced the gold
from the full context, and none of the five constant answers was accepted.
Whether each is a constraint question is part of the human check
(Appendix~\ref{app:human-val}). \cat{Aspect} questions name 2.6 lenses on
average (range 2--4).
 
\begin{table}[h]
\centering
\small
\setlength{\tabcolsep}{4pt}
\begin{tabular}{@{}lrrrr@{}}
\toprule
Category & $n$ & Evidence & Sessions & Span \\
\midrule
\cat{Single-hop}  & 168 & 1.11 & 1.05 & 0.08 \\
\cat{Multi-hop}   & 76  & 2.32 & 1.54 & 0.83 \\
\cat{Temporal}    & 74  & 2.01 & 1.95 & 1.78 \\
\cat{Open-domain} & 73  & 5.07 & 3.25 & 3.29 \\
\midrule
\cat{Revision}    & 64  & 4.25 & 3.55 & 9.03 \\
\cat{Stance}      & 50  & 3.90 & 2.60 & 5.36 \\
\cat{Constraint}  & 148 & 2.28 & 1.82 & 4.86 \\
\cat{Aspect}      & 39  & 4.36 & 2.44 & 5.92 \\
\bottomrule
\end{tabular}
\caption{Mean evidence turns, sessions cited and session span per
question, by category.}
\label{tab:question-stats}
\end{table}
 
\paragraph{Question filtering.}
Standard questions are generated with LoCoMo's pipeline. Of its 464
questions, 391 are scored for accuracy; the other 73 are unanswerable and
are scored separately, as abstention. A pass over numeric answers that
appeared to contradict another mention judged 29 candidates and removed 1.
 
Modes candidates are written by MiniMax-M3 and kept only if they pass four
checks, in order. The \emph{structural} check enforces the category:
evidence ids exist, exactly one speaker is named, and the question does
not pin a date; a \cat{Revision} question lists values in at least two
sessions and is either out of order or has at least three entries, where a
repeated mention of the same number counts as an entry; a \cat{Constraint}
question shares at most a third of its content words with the rule; and
an \cat{Aspect} question cites at least two lenses. The \emph{blind} check
gives MiniMax-M3 the full context but not the gold, and the same model
must judge its answer correct. The \emph{chance-floor} check judges five
constant answers without any conversation and rejects the candidate if
any of them is marked correct. Finally, duplicates of earlier questions
are dropped. Because the generator and the judge are the same model, these
filters do not replace the human check (Appendix~\ref{app:human-val}). The
blind check also ensures that every Modes question is answerable from the
full context, which favours the full-context reference.
 
Table~\ref{tab:funnel} reports the filters on an earlier build; about a
third of its candidates were kept. The released Modes set consists of the
301 questions accumulated over later runs of the same filters.
\cat{Aspect} loses most candidates to the structural check, mainly for
covering fewer than two sessions or two lenses, which is why it is the
smallest category.
 
\begin{table}[h]
\centering
\small
\setlength{\tabcolsep}{4pt}
\begin{tabular}{@{}lrrrr@{}}
\toprule
Category & Generated & Kept & Blind & Other \\
\midrule
\cat{Revision}   & 78  & 38  & 23  & 17 \\
\cat{Stance}     & 72  & 33  & 31  & 8  \\
\cat{Constraint} & 71  & 26  & 41  & 4  \\
\cat{Aspect}     & 74  & 9   & 22  & 43 \\
\midrule
All              & 295 & 106 & 117 & 72 \\
\bottomrule
\end{tabular}
\caption{Candidates removed by each filter in an earlier build of the
Modes split. \emph{Blind}: failed the blind check. \emph{Other}: failed
the structural, chance-floor or duplicate check.}
\label{tab:funnel}
\end{table}

\subsection{Mode Split Examples}\label{app:examples}
Table~\ref{tab:modes-examples} shows one example of each question
type from the Modes split.

\begin{table*}[t]
\centering
\small
\renewcommand{\arraystretch}{1.15}
\begin{tabularx}{\textwidth}{@{}p{1.45cm}p{3.3cm}Xp{3.6cm}p{2.3cm}@{}}
\toprule
\textbf{Type} & \textbf{Question} & \textbf{Evidence} & \textbf{Retrieval trap} & \textbf{Gold} \\
\midrule
\cat{Revision} &
What discount rate is Sarah currently using for her SPG DCF? &
\textbf{D16:3} ``Using 8.0\% for SPG now, up from 7.75\%.'' \newline
\textbf{D17:2} ``10Y near 4.2\% pushes my discount rate to 8.25\%.'' &
\textbf{D17:6}, the latest mention: ``at 4.3\% I'd bump SPG's discount rate up another 25bps'' (hypothetical) &
8.25\% \\
\midrule
\cat{Stance} &
What data point caused Robert to first raise his XOM price target above \$126? &
\textbf{D21:3} ``\$126 target still works if Brent holds here.'' \newline
\textbf{D22:3} ``\$94 Brent has me second-guessing my mid-cycle call \ldots\ XOM's \$126 target needs a \$5--10 bump.'' \newline
\textbf{D23:5} ``updated my PT to \$131 from \$126.'' &
\textbf{D23:5} states the raise but not its cause; \textbf{D25:3} reports a later raise to \$136 &
Brent reaching \$94, prompting a late-cycle rethink \\
\midrule
\cat{Constraint} &
Given Kenji's trailing-stop discipline, at what price would his stop trigger on the Sony position he reloaded at \textyen13,420? &
\textbf{D21:5} ``30bps trailing, same as before \ldots\ not giving back more than 0.3\% on any single name.'' \newline
\textbf{D26:5} ``picked up 1\% Sony last Wednesday at \textyen13,420.'' &
\textbf{D26:5} matches the question but carries no rule; the rule is five sessions earlier &
$\approx$\textyen13,380 \\
\midrule
\cat{Aspect} &
How does Eleanor reconcile liking SPG's valuation with her reluctance to size it up, and what risk lens dominates her hesitation? &
\textbf{D25:4} ``SPG feels like the steadier play---100 shares there is sleep-at-night money.'' \newline
\textbf{D26:4} ``Same CEO across two 90\%+ payout names is a concentration risk red flag.'' \newline
\textbf{D26:10} ``my 100 SPG shares at 6.65\% yield still feel like the right anchor.'' &
\textbf{D22:5} cites a different risk: ``100 SPG shares funding that 6.65\% yield with debt has me nervous'' &
Likes the 6.65\% yield but holds at 100 shares; same-CEO concentration risk \\
\bottomrule
\end{tabularx}
\captionof{table}{Example questions from the Modes split. \emph{Evidence} lists the
turns (session:turn) the gold depends on; \emph{Retrieval trap} is a turn
that a similarity- or recency-based retriever is likely to return instead.}
\label{tab:modes-examples}
\end{table*}

\section{\sys{} Details}
\subsection{Topic taxonomy}
\label{app:taxonomy}
 
Views are keyed by an L2 topic and tagged with L3 topics from a fixed three-level taxonomy: 5 analytical domains (L1), 20
subtopics (L2) and 146 leaf tags (L3), each leaf with aliases for matching. Table~\ref{tab:taxonomy} lists the first two levels.
 
\begin{table}[h]
\centering
\small
\begin{tabular}{@{}p{2.2cm}p{4.9cm}@{}}
\toprule
L1 domain & L2 subtopics \\
\midrule
Fundamental analysis & Income statement, Balance sheet, Cash flow, Valuation, Management quality \\
Market dynamics & Price action, Institutional flow, Sentiment, Liquidity \\
Corporate actions & Earnings events, M\&A activity, Capital structure, Management changes \\
Quantitative \& risk & Factor exposure, Portfolio risk, Earnings quality \\
Macro \& sector & Macro environment, Sector dynamics, Geopolitical risk, Commodities \& FX \\
\bottomrule
\end{tabular}
\caption{L1 and L2 levels of the topic taxonomy.}
\label{tab:taxonomy}
\end{table}

\subsection{Memory Representation and Construction Details}\label{app:rep-and-cons}

\paragraph{Figure kinds.} Commitments: position cap or size, add or
trim level, price target, multiple, yield, estimate. Observations:
market level, third-party estimate, executed fill, P\&L.

\paragraph{Anchor fields.} Each anchor also stores a canonical unit, a
comparator (at, below, above), a short description of what it measures,
and the speaker the figure belongs to.

\paragraph{Attribute kinds.} Risk tolerance, risk focus, goal,
constraint, time horizon, experience and preference.

\paragraph{Claim fields.} Each claim stores its event time, the speaker,
and the entities and topic it concerns. Claims can also hold assertions
about things that are not stored entities, which views and anchors
cannot.

\paragraph{Event time versus processing time.} The two diverge when
sessions are ingested in bulk. Every node also records the identifiers
of the turns it was derived from.
 
\paragraph{Extraction.}
One schema-constrained call per session returns entities, (entity, topic)
pairs, anchors, claims and attributes from the user's side of the
conversation. Output that breaks a structural rule of the schema (an entity
not linked to the user or without a topic, an unknown attribute kind, a
relation to an undeclared entity) is rejected and the call retried, up to
five times. Entity identifiers are
folded onto the spellings already stored in $\mathcal{G}_p$.
Table~\ref{tab:vocab} lists the closed vocabularies.
 
\begin{table}[h]
\centering
\small
\begin{tabular}{@{}p{1.9cm}p{5.2cm}@{}}
\toprule
Field & Values \\
\midrule
Anchor kind, commitment & position cap, position size, add level, trim level, price target, valuation multiple, yield, estimate \\
Anchor kind, observation & observed level, third-party estimate, executed fill, P\&L \\
Unit & currency, percent (basis points are normalised to percent), multiple, shares, quantity \\
Comparator & at, below, above \\
Source & user, endorsed (the other party's figure, adopted by the user), counterparty, third party \\
Action kind & initiated, added, trimmed, exited, upgraded, downgraded, target raised, target cut, stop tightened, stop loosened, hedged, deferred, held \\
Attribute kind & risk tolerance, risk focus, goal, constraint, time horizon, experience, preference \\
Sentiment & bullish, bearish, neutral, mixed \\
\bottomrule
\end{tabular}
\caption{Closed vocabularies of the extraction schema.}
\label{tab:vocab}
\end{table}
 
\paragraph{Stored fields.}
Besides its key, each anchor stores a value in a canonical unit, a
comparator (at, below, above), a short description of what it measures,
and whose figure it is (Table~\ref{tab:vocab}). For yields and estimates
the period is the reporting period, so values for different periods
coexist. The topic is not part of the anchor key, so a commitment
discussed under two lenses remains one quantity. A view's state is its
text and sentiment, an anchor's its value, unit, currency and comparator,
and an attribute's its text and level; a node is snapshotted only when
this state changes, and a session already applied to a node is ignored.
 
\paragraph{Quantity resolution.}
For each new commitment $a$, the candidates $Q(a)$ are the stored anchors
with the same entity, kind, unit, currency and period; an anchor in another
unit is never a candidate, since a percentage cannot revise a price. Each
anchor carries a short description of what it measures, written by the
extractor. The resolver sees the new figure's description and the
candidates' descriptions and current values, and decides for each new figure
whether it is the same quantity as one candidate. It is instructed that a
figure is a different quantity, even with the same entity, kind and unit,
when it concerns another account or portfolio sleeve, another scenario,
planned rather than actual sizing, a part rather than the whole (the amount
trimmed rather than the resulting position), a different trigger, instrument
or horizon, or somebody else's number, and to answer \emph{different} when
the descriptions do not settle it. A match adopts the stored slot $j$, so the
write supersedes it; otherwise $a$ receives a new slot derived from its
description. Observations are not resolved, since their period is the date
on which they were stated.
 
\paragraph{Synthesis.}
After the update, the text of every view and attribute the session touched
is re-synthesised from its previous text and the session, at temperature 0.
At 0.2, re-ingesting identical data moved one speaker's accuracy from 71\%
to 60\%.

\subsection{Retrieval configuration}
\label{app:config}
 
\paragraph{Routing.}
The router returns the question's entities $E_q$, topics $Z_q$, a type
(stance, quantity, temporal, profile, exploratory or comparative), whether
it concerns past states, an optional reference date, and a grounding:
entity-anchored, topic-anchored or semantic-only. Entities are resolved to
identifiers in $\mathcal{G}_p$ before any search, by exact identifier,
alias, lexical match and then embedding match; the user's own name, and
other people's, are excluded from resolution. In the evaluation, the
graphs queried are those of the speakers the question names, or both
when it names neither.
 
\paragraph{View scoring.}
Current views are ranked by
\begin{equation*}
\begin{aligned}
s(v) &= \cos(\mathbf{q},\mathbf{v})
+ \lambda_e\,\mathbb{1}[e\in E_q]
+ \lambda_z\,\mathbb{1}[z\in Z_q]\\
&\quad + \lambda_t\,\mathbb{1}[\mathrm{tags}(v)\cap Z_q\neq\varnothing],
\end{aligned}
\end{equation*}
with each indicator binary and $(\lambda_e,\lambda_z,\lambda_t)$ set by
hand for each grounding (Table~\ref{tab:weights}). The search returns at
most $\max(3k,12)$ candidates with $k=8$; a view is admitted only if its
raw cosine similarity is at least $\theta=0.70$, so that the bonuses can
reorder candidates but not admit an unrelated one, at most two views per
entity are kept, and the top $k$ remain. Two expansions then append up to
five views of entities one hop away in the shared entity graph (related,
competitor, part-of), ranked by similarity, and up to five views tagged
with $Z_q$, ranked by recency; these are not filtered by $\theta$.
 
\begin{table}[h]
\centering
\small
\begin{tabular}{@{}lccc@{}}
\toprule
Grounding & $\lambda_e$ & $\lambda_z$ & $\lambda_t$ \\
\midrule
Entity-anchored & 0.10 & 0.02 & 0.02 \\
Topic-anchored  & 0.03 & 0.08 & 0.05 \\
Semantic-only   & 0.02 & 0.02 & 0.01 \\
None returned   & 0.05 & 0.03 & 0.02 \\
\bottomrule
\end{tabular}
\caption{Weights of the entity, topic and tag terms of the view score, by
the question's grounding. The values are set by hand.}
\label{tab:weights}
\end{table}
 
\paragraph{Channels by question type.}
\begin{itemize}[leftmargin=*,noitemsep,topsep=2pt]
  \item \emph{Anchors}: all kinds for quantity, profile and exploratory
  questions; for stance and temporal questions, only the decision kinds
  (position size and cap, executed fill, add and trim levels, price target).
  Each anchor is returned with its value history.
  \item \emph{Past views}: for every question except present-tense stance
  and profile questions, ranked by similarity plus the entity term,
  restricted to states valid on the reference date when one is given; up
  to five.
  \item \emph{Claims}: ranked by similarity over all of the user's claims,
  with the floor $\theta$; up to eight. Their source turns enter $U$.
  \item \emph{Trajectory}: for stance and temporal questions and questions
  about the past, the user's claims about the resolved entities (or, for a
  theme, about $Z_q$), in event order; up to 24, keeping actions first
  when thinned.
  \item \emph{Attributes}: always.
\end{itemize}
 
\paragraph{Grounding check.}
A question is marked ungrounded when no view reaches $\theta$, or when none
of the entities it names (other than people) resolves in $\mathcal{G}_p$ and
none of its topics matches a retrieved view. If, in addition, no anchor is
retrieved, the structured sections are replaced by a notice that nothing in
memory supports an answer; the verbatim turns are still appended.
 
\paragraph{Context assembly.}
When the question names known entities, their views lead and at most two
views about other entities are kept. Requested anchor kinds are listed
first. The verbatim section $U$ holds up to 20 source turns of the retrieved
claims and trajectory, deduplicated across the speakers queried, plus the
15 turns most similar to the question among the remaining source turns of
all retrieved views, past views, claims and trajectory points; all are
shown in chronological order. The claims themselves are not rendered. On Modes this gives about 27 turns
and 2.6k tokens per question.

\subsection{Deployed pipeline}\label{app:deployment}
\begin{figure*}[h]
  \centering
  \includegraphics[width=0.85\textwidth]{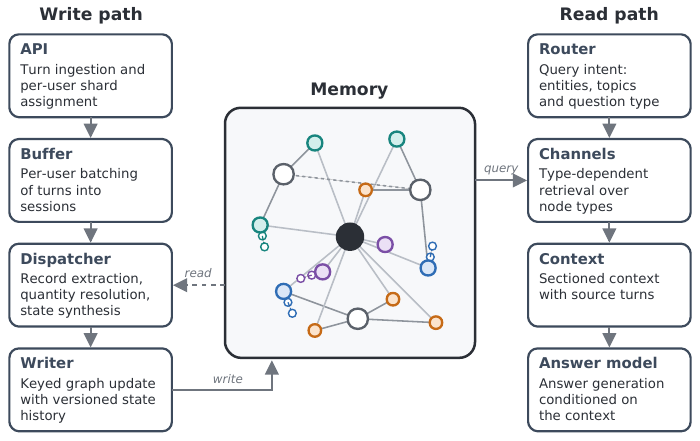}
  \caption{Write and read paths of the deployed \sys{}.}
  \label{fig:pipeline}
\end{figure*}

Figure~\ref{fig:pipeline} shows the write and read paths of the deployed
system.
 
\section{Additional results}
\label{app:results}
\subsection{Equal retrieval budget.}
Table~\ref{tab:budget} gives similarity retrieval over turns the same
number of turns as \sys{}. Retrieval is ahead only on \cat{stance}, whose
evidence often lies in turns that do not name the entity. At 42 turns it
reaches recall 0.88, and the lead of \sys{} narrows to 2.2 macro and 10.3
on \cat{revision}.
\begin{table}[t]
\centering
\small
\setlength{\tabcolsep}{7pt}
\begin{tabular}{@{}lccc@{}}
\toprule
 & Turn RAG & \sys{} & $\Delta$ \\
\midrule
\cat{revision} & 70.2 & \textbf{86.5} & \up{16.3} \\
\cat{stance}   & \textbf{65.3} & 61.5 & \dn{3.8} \\
\cat{Constraint}   & 74.8 & \textbf{79.1} & \up{4.3} \\
\cat{aspect}         & 63.2 & \textbf{68.4} & \up{5.2} \\
\midrule
Macro        & 68.4 & \textbf{73.9} & \up{5.5} \\
Recall       & 0.82 & 0.80 & \\
\bottomrule
\end{tabular}
\caption{Equal retrieval budget on Modes: similarity retrieval over turns
(no graph) and \sys{}, each showing about 27 turns.}
\label{tab:budget}
\end{table}

\subsection{Abstention.}
Table~\ref{tab:abstain} scores each reader's declines over all 765
questions, with the 73 unanswerable questions as positives.

% Abstention 
\begin{table}[t]
\centering
\footnotesize
\setlength{\tabcolsep}{3.6pt}
\begin{tabular}{@{}l ccc ccc@{}}
\toprule
 & \multicolumn{3}{c}{Gemini} & \multicolumn{3}{c}{Nemotron} \\
\cmidrule(lr){2-4}\cmidrule(l){5-7}
Method & Prec. & Rec. & F1 & Prec. & Rec. & F1 \\
\midrule
Mem0            & 26.0 & \textbf{93.2} & 40.6 & 23.8 & \textbf{98.6} & 38.4 \\
A-Mem           & 69.1 & 89.0 & 77.8 & 59.2 & 83.6 & 69.3 \\
MemGPT          & 81.3 & 83.6 & \textbf{82.4} & 80.0 & 82.2 & \textbf{81.1} \\
Graphiti (Zep)  & 19.0 & 71.2 & 30.0 & 15.5 & 82.2 & 26.1 \\
O-Mem           & 22.5 & 52.1 & 31.4 & 21.9 & 57.5 & 31.7 \\
\sys{}          & \textbf{91.5} & 74.0 & 81.8 & \textbf{87.8} & 58.9 & 70.5 \\
\midrule
\emph{Full context} & 73.3 & 75.3 & 74.3 & 86.4 & 52.1 & 65.0 \\
\bottomrule
\end{tabular}
\caption{Precision and recall of a decline. A true positive is a decline on one
of the 73 unanswerable questions, counted over all 765 questions.}
\label{tab:abstain}
\end{table}

\section{Prompts}
\label{app:prompts}
 
This section gives the instruction text of every model call in \sys{} and
in the evaluation. Angle-bracket placeholders (\texttt{<QUESTION>}) and
Python format fields (\texttt{\{question\}}) are filled at run time; they are
not further instructions. The prompts are reproduced verbatim except that
the embedded taxonomy of the extraction prompt is replaced by the
placeholder \texttt{<TOPIC\_TAXONOMY\_JSON>} (its ids are listed in the
router prompt, \S\ref{app:prompt-router}, and its first two levels in
Table~\ref{tab:taxonomy}), one example file name is anonymised, and
non-ASCII punctuation is transliterated. Every call returns JSON validated
against a schema; we list the output fields here, and the full schemas are
in the released code. Table~\ref{tab:prompt-calls} summarises the calls.
 
\begin{table}[t]
\centering
\footnotesize
\setlength{\tabcolsep}{3pt}
\begin{tabular}{@{}l l r >{\raggedright\arraybackslash}p{2.4cm}@{}}
\toprule
Call & Model & Max & Output fields \\
\midrule
Extraction      & Gemini     & 65{,}536 & entities, relations, anchors, claims, attributes \\
Synthesis       & Gemini     & 8{,}192  & text and sentiment per (entity, topic) row \\
Resolver $\psi$ & Gemini     & 1{,}024  & stored quantity revised, or none \\
Router          & Gemini     & 1{,}024  & entities, topics, type, date, grounding \\
\midrule
Reader          & Gemini     & 1{,}024  & value, answer, declined, support \\
Reader          & Nemotron   & 8{,}192  & as above \\
Judge           & MiniMax-M3 & 256      & verdict, reason \\
\bottomrule
\end{tabular}
\caption{Model calls, all at temperature 0. \emph{Max}: maximum output
tokens. Gemini is \texttt{gemini-3.1-flash-lite} and Nemotron is
\texttt{nemotron-3.5-lightning}. The first four calls build or query the
memory of \sys{}
(Appendix~\ref{app:prompt-extract}--\ref{app:prompt-router}); the reader
and judge are shared by all methods. The resolver runs only when a new
commitment has stored candidates.}
\label{tab:prompt-calls}
\end{table}

\subsection{Extraction}
\label{app:prompt-extract}
One call per session (\textsc{Extract} in Algorithm~\ref{alg:write}). \texttt{<USER\_ID>}
is the graph owner; turns are numbered in order and labelled \texttt{[user]}
for the speaker being ingested and \texttt{[assistant]} for the other
participant. The call classifies entities and relations and extracts
anchors, claims and attributes; it writes no prose, which is left to
synthesis.
 
\Needspace{8\baselineskip}
\begin{lstlisting}[style=prompt,title={Extraction prompt}]
You are a financial knowledge graph extraction specialist.
Your job: classify entities and relationships only. Do NOT write context sentences.
 
USER: <NAME>   (graph id: <USER_ID> -- emit this exact id as the USER identifier)
CONVERSATION:
<NUMBERED_CONVERSATION_TURNS>
 
========================================================
STEP 1 -- ENTITY EXTRACTION
========================================================
Identify every meaningful real-world entity the user or assistant explicitly discusses.
 
NODE TYPES:
  USER   -- emit exactly one USER node: identifier="<USER_ID>", label="<USER_ID>".
  ENTITY -- any real-world thing: equity, ETF, person, company, product, index, fund,
           concept, event, metric, strategy, etc.
           Identifiers: uppercase ticker for equities (e.g. "NVDA"), snake_case for all else.
           Always set entity_subtype: TICKER, ETF, INDEX, FUND, COMPANY, PERSON,
           ORGANIZATION, EVENT, METRIC, STRATEGY, CONCEPT, or a short snake_case descriptor.
 
ENTITY RULES:
  0. IT MUST BE A SUBJECT, NOT AN ARTEFACT OF THIS CONVERSATION.
     Rule 1 asks whose entity it is. This asks whether it is an entity at all, and it is
     checked FIRST -- engagement cannot rescue something that should never have been a node.
 
     THE TEST: could <USER_ID> still hold a view about this a year from now, in a
     conversation that has nothing to do with today's? A company, a ticker, a sector, a
     macro theme, a strategy, a person in the market -- yes. The file they just uploaded,
     the document being drafted, the tool they are using -- no. Those exist only because
     this session happened.
 
     NEVER emit an entity for:
       - FILES AND DOCUMENTS in the workspace -- an uploaded PDF, a deck, a spreadsheet, a
         report being written, a guide, a note. A document is a SOURCE, not a subject. The
         companies discussed inside it are the entities; the document itself is not.
         (Observed in production: "<file name>", "guide_1".)
       - THE SOFTWARE AND ITS PARTS -- this product, its panes, its agent, its database, its
         models or pipelines, and comparisons between them and anything else. A user
         discussing how the tool works is not expressing a market view.
         (Observed in production: "financial_platform", "sell_side_model".)
       - THE SESSION ITSELF -- a task, a request, a chart just produced, an error, a
         schedule, "the report", "the analysis".
       - GENERIC ABSTRACTIONS with no referent a market participant would recognise, e.g.
         "the platform", "the workflow", "data quality" as a bare notion.
 
     A conversation may legitimately produce NO entities. Engineering discussion, product
     feedback, tooling questions and workspace admin are normal uses of this system and
     they are not investment memory. Emitting nothing is the correct output for them --
     far better than forcing a fit, which stores a view <USER_ID> does not hold about a
     thing that is not an entity.
 
     Fictional or private companies ARE admissible when the conversation treats them as
     real subjects of analysis (a synthetic portfolio, a private credit borrower). The test
     is whether it is a market subject, not whether you can look it up.
 
  1. THE USER MUST HAVE ENGAGED WITH IT. This memory records <USER_ID>, not the assistant.
     Extract an entity only when <USER_ID> either:
       (a) names it themselves -- "I hold KO", "what about TSLA at these levels", or
       (b) RESPONDS to something the assistant raised -- in EITHER direction. Judge this
           from meaning, not keywords, and note the reply often never repeats the name:
             accepting  -- "I agree", "good suggestion", "that one I like",
                          "makes sense, I'll add some", "let me look at that"
             rejecting  -- "no thanks", "I don't want energy exposure", "not for me",
                          "I've looked at it before and passed"
           A REJECTION IS ENGAGEMENT, and often the more valuable memory: it records a
           view <USER_ID> actually holds. Capture it as INTERESTED_IN with the negative
           sentiment, not as a drop.
           Where a bare reply ("I agree") could refer to several entities the assistant
           named, apply it to the one(s) it most plausibly answers; if that is genuinely
           undecidable, leave them all out.
     An entity the assistant raised that <USER_ID> passed over, changed the subject
     from, or simply did not respond to is NOT extracted at all -- not as MENTIONED,
     not with low confidence, not at all.
 
     The two cases look alike and are opposite. Compare them carefully:
       NO RESPONSE -> drop
         Assistant: "income investors also love XOM right now."
         <USER_ID>:  "Within staples I'd stick with KO and PG."
         -> extract KO and PG. Do NOT extract XOM: <USER_ID> moved past it without
           evaluating it.
       DECLINED -> extract, with the negative view
         Assistant: "some investors allocate 1-2% to BTC as 'digital gold'."
         <USER_ID>:  "Not for me -- I don't touch crypto, no cash flows to value."
         -> DO extract BTC as INTERESTED_IN / BEARISH. <USER_ID> evaluated it and said
           no, and gave a reason. That is one of the most durable memories available:
           it predicts every future crypto suggestion. Dropping it is a serious loss.
     The test is not whether <USER_ID> was positive -- it is whether <USER_ID> ENGAGED.
     When it is genuinely unclear whether the user took it up, leave it out. A missing
     memory is recoverable from a later conversation; a memory recording an interest
     <USER_ID> never had is not.
 
     APPLY THIS AS A CHECK, NOT A PRINCIPLE. Before emitting ANY entity, find the
     specific [user] turn that engages it and be able to quote it. If the only turns
     naming it are [assistant] turns, DROP IT -- no matter how substantive the assistant
     was, how central it seems to the discussion, or how naturally it belongs beside
     entities you are keeping. Two entities named in one assistant sentence, where the
     user replies about only one, means exactly one is extracted.
       [assistant] "I'll have the soy weather bullet ready, since that USDA report
                    could shake things up."
       [user]      "Sounds good -- talk Friday."
       -> extract NEITHER soy NOR the USDA report. A scheduling reply is not engagement
         with the subjects mentioned around it.
  2. Prefer the most specific form: "msft_cloud_segment" over "cloud".
  3. Skip: pronouns, bare time words (today, quarter), filler nouns (stuff, things).
  4. SKIP WORK PRODUCTS AND LOGISTICS. A thing <USER_ID> will WRITE, SEND or DO is not a
     thing they hold a view about, even though they named it themselves:
       notes, memos, recaps, decks, models, spreadsheets, reports they are producing;
       meetings, calls, deadlines, calendar items.
       "I'll get the Meridian tightness note and junior assay recap done by Friday"
       -> extract NOTHING from this. It records a task, not a stance.
     The subject a document is ABOUT may still be an entity if <USER_ID> says something
     substantive about it -- but the document itself never is.
  5. Each entity appears ONCE in the entities list.
 
========================================================
STEP 2 -- RELATIONS
========================================================
Classify the relationship between the user and each entity.
 
USER->ENTITY relation types:
  MENTIONED     -- <USER_ID> engaged with the entity (named it, or took up the assistant's
                  mention of it) but revealed no directional view of their own. Use this for
                  a name dropped in passing, a neutral follow-up question, or a bare
                  acknowledgement.
                  NB an entity <USER_ID> never took up is not MENTIONED -- it is not
                  extracted at all (ENTITY RULE 1).
  INTERESTED_IN -- the user holds a thesis, interest, or directional stance -- expressed
                  directly, OR revealed by endorsing, building on, or acting on a point
                  the assistant raised.
                  sentiment: REQUIRED -- BULLISH / BEARISH / NEUTRAL / MIXED.
 
ENTITY->ENTITY relation types:
  RELATED_TO    -- two entities are meaningfully linked in this conversation.
  COMPETES_WITH -- two entities compete in the same market.
  PART_OF       -- DIRECTION MATTERS: subject is the PART, object is the WHOLE.
                  "grasberg PART_OF FCX" -- the mine belongs to the company, never the
                  reverse. Same for a segment in a company, a holding in a fund, a
                  constituent in an index.
 
RULES:
  1. Every ENTITY must appear in at least one USER->ENTITY relation. No orphans.
  2. Only emit INTERESTED_IN when the USER reveals the view -- directly, or by endorsing,
     building on, or acting on a point the assistant raised. Facts and framing may come
     from the assistant, but the STANCE must be the user's.
  3. An assistant assertion the user merely heard and did not take up is NOT a user view.
     If the user took it up but revealed no direction, that is MENTIONED; if they did not
     take it up at all, the entity is dropped entirely (ENTITY RULE 1).
  4. Do not emit HAS_SUBTOPIC.
  5. Do NOT write context sentences -- that happens in a separate step.
 
========================================================
STEP 3 -- TOPIC TAGGING  (MANDATORY for every ENTITY)
========================================================
Every user-connected ENTITY must have at least one TAGGED_WITH relation.
Tags capture WHY the user engaged with this entity -- they will be attached
to the user's memory, not to the entity itself.
 
RULES:
  1. Tag based on the aspect the user discussed, not what the entity is in general.
     GOOD: NVDA -[TAGGED_WITH]-> l3_eps          (user discussed NVDA's earnings)
     BAD:  NVDA -[TAGGED_WITH]-> l1_fundamental  (too broad -- pick the specific angle)
  2. Persons and concepts must be tagged too -- use the angle the user discussed:
     GOOD: jensen_huang          -[TAGGED_WITH]-> l3_track_record   (CEO credibility)
     GOOD: nvda_datacenter_seg   -[TAGGED_WITH]-> l3_revenue_growth (segment growth)
     GOOD: china_export_controls -[TAGGED_WITH]-> l3_export_controls
  3. One entity may have multiple tags if the user discussed multiple distinct angles.
  4. Prefer TOPIC_L3 leaves; fall back to L2, then L1.
  5. Emit TOPIC_* nodes (identifier = exact taxonomy id) in the entities list.
  6. Emit TAGGED_WITH edges: ENTITY -[TAGGED_WITH]-> TOPIC_*. No context.
  7. Only use ids from the taxonomy below -- do NOT invent new ids.
  8. An entity with no TAGGED_WITH relation is an extraction error.
 
TOPIC TAXONOMY:
<TOPIC_TAXONOMY_JSON>
 
========================================================
STEP 3b -- TYPED ANCHORS  (every concrete figure in the conversation)
========================================================
Attach to a USER->ENTITY relation EVERY specific number stated in this conversation -- both
what <USER_ID> committed to AND what was merely observed or cited. Stored as typed fields,
a figure can be looked up exactly instead of competing for room inside a prose sentence
that has a hard character budget. THE PROSE IS NOT A SAFETY NET: it is written later, to a
budget, by a different call, and figures left to it are routinely lost.
 
WORK FIGURE BY FIGURE. Scan <USER_ID>'s turns for every number -- prices, levels, sizes,
percentages, counts, dates-as-levels -- and for each one ask which kind fits. Do not stop
after the first two or three. A conversation with eight figures should usually yield
several anchors, not one.
 
KINDS -- use exactly one of these; if a number genuinely fits none, leave it out.
 
COMMITMENT -- what <USER_ID> DECIDED. A later value replaces the earlier one:
  POSITION_CAP   -- max weight they will hold        ("I cap NVDA at 7%")
  POSITION_SIZE  -- current weight                    ("TSM is 4% of the book")
  ADD_LEVEL      -- level/condition to buy more       ("adding on any dip below 850")
  TRIM_LEVEL     -- level/condition to trim or stop   ("I'd trim above 10%")
  PRICE_TARGET   -- their fair-value or target price  ("I see fair value at 195")
  VALUATION_MULTIPLE -- qualifier REQUIRED: P/E, EV/EBITDA, EV/S  ("90x earnings")
  YIELD          -- period-indexed                    ("BND yielded 4.2% last year")
  ESTIMATE       -- qualifier + period REQUIRED       ("peak GLP-1 revenue of $20B in FY2030")
 
OBSERVATION -- what was TRUE at a moment, or what somebody else asserted. These accumulate
by date instead of replacing each other, so two readings taken weeks apart both stand:
  OBSERVED_LEVEL -- ANY market datum as of now, not just price:
                     price      ("copper just hit $8,510", "10y at 3.52%")
                     inventory  ("LME stocks at 151,750 tonnes")
                     spread     ("LME-Comex dislocation of $80/tonne")
                     volume, and period-over-period CHANGES in any of them
                                ("stocks down 6.5% week-over-week" -> value 6.5 PCT,
                                 qualifier "LME stocks WoW change")
                   Emit one anchor PER FIGURE: "stocks at 151,750 tonnes, down 6.5% WoW"
                   is TWO -- the level and the change.
  THIRD_PARTY_ESTIMATE -- someone else's number      ("JPM cut CCL EBITDA 18%",
                                                     "LLY guided Mounjaro to $4.2B")
  EXECUTED_FILL  -- a trade actually done, or cost basis
                                                    ("1,200 shares at $38 cost basis",
                                                     "entered EUR/USD at 1.0820")
  PNL            -- realised or mark-to-market result ("book down 14.5% YTD",
                                                     "locked in that 12%")
 
FIELDS: value (unscaled -- $25B is 25000000000), unit (CURRENCY/PCT/BPS/MULTIPLE/SHARES/
QUANTITY -- use CURRENCY for money and set currency to the ISO code, defaulting to USD; say
BPS if that is how it was phrased and it will be stored as PCT; use QUANTITY for a counted
or measured amount and put the unit of measure in the qualifier: "151,750 tonnes" is
value 151750, unit QUANTITY, qualifier "tonnes"),
comparator (AT / BELOW / ABOVE -- "below 850" is BELOW, "target 195" is AT),
qualifier, period, source (USER / ENDORSED),
basis -- what the number MEASURES, in ~12 words, specific enough to tell it from another
figure of the same kind on the same entity: "AMD weight in retiree accounts", "hard stop on
the half-size re-entry", "discount rate in the SPG base-case DCF". Name the account,
scenario (base / downside), leg or trigger whenever the conversation does. ALWAYS fill it:
a later session decides from this whether its figure REVISES this one or is a different
quantity, and "position size" alone cannot answer that.
 
SCALE -- write the number exactly as it was said, never converted:
  "8%"      -> value 8.0    unit PCT        NOT 0.08. A percent is the number spoken.
  "0.5%"    -> value 0.5    unit PCT        (sub-1% values are legitimate and rare)
  "25 bps"  -> value 25     unit BPS        (stored as 0.25 PCT automatically)
  "$25B"    -> value 25000000000  unit CURRENCY currency USD
  "40x"     -> value 40     unit MULTIPLE
  Getting this wrong is not a rounding error: a cap stored as 0.08 and later restated as
  8.0 looks like a 100x revision, which destroys the supersession the anchor exists for.
 
RULES:
  1. PICK THE TIER FIRST, THEN THE KIND. The test is whose number it is and whether it can
     be revised:
       <USER_ID> decided it, and could change their mind  -> COMMITMENT kind
       it was true at a moment, or somebody else said it  -> OBSERVATION kind
     "I cap NVDA at 7%" is a commitment. "Copper hit $8,510" is an observation -- it will
     never be *revised*, it will simply be a different number tomorrow.
     A general truth with no number attached to a moment or a decision -- "3nm is sold out
     through 2026", "the Fed cut twice" -- is still NOT an anchor; it belongs in the context
     sentence. A wrong anchor is worse than a missing one, because it is trusted as
     structured truth.
  2. SOURCE IS MANDATORY AND CARRIES THE ATTRIBUTION:
       USER         -- <USER_ID> stated it as their own
       ENDORSED     -- the other party said it and <USER_ID> took it up ("I agree")
       COUNTERPARTY -- the other party said it and <USER_ID> did NOT adopt it
       THIRD_PARTY  -- attributed to a bank, company guidance, or an index
     Getting this wrong is how somebody else's number gets read back as <USER_ID>'s view.
     Most OBSERVATION anchors are COUNTERPARTY or THIRD_PARTY, not USER.
  3. Only extract a figure that is actually present. NEVER infer, round, or compute one.
  4. period ONLY for YIELD and ESTIMATE -- the quantities that take different values at
     different reporting dates. Two such values for DIFFERENT periods both stand; for the
     SAME period the later one replaces the earlier.
     Leave period NULL on OBSERVATION kinds unless the figure names an explicit reporting
     period: they are dated by the conversation automatically.
  5. Emit nothing when the conversation contains no such quantity. Most turns will not.
  6. AN ANCHOR IS ABOUT ONE ENTITY. A rule covering the whole book is NOT an anchor --
     it is a CONSTRAINT self-attribute (STEP 4), stored once against <USER_ID>.
       "nothing above 8%"        -> CONSTRAINT (a portfolio rule), NOT a POSITION_CAP on
                                   every name discussed
       "I keep 10% in cash"      -> CONSTRAINT
       "I cap NVDA at 7%"        -> POSITION_CAP anchor on NVDA (names one entity)
     Test: if you are about to attach the SAME anchor to two or more entities, it is a
     portfolio rule -- emit the CONSTRAINT once instead. Copying it onto each name means a
     later change reaches only the names discussed that day and the rest silently keep the
     old value, which is the drift anchors exist to prevent.
 
========================================================
STEP 3c -- CLAIMS  (the fine-grained record)
========================================================
Emit one CLAIM per substantive assertion in this conversation. Steps 1-3 build a
structured INDEX; this is the record the index points at.
 
A claim is one sentence, self-contained, naming its subject rather than saying "she" or
"it", with every figure kept verbatim. Set exchange_id to the turn identifier shown in
the conversation.
 
WHAT TO EMIT -- anything either party actually asserted:
  a position, level, target, threshold, or rule      ("<NAME> caps single names at 7%")
  a view and its reason                              ("<NAME> likes SCHN at 22x for grid exposure")
  an action taken                                    ("<NAME> trimmed Orsted by 50bps on China lockdowns")
  a condition or commitment                          ("<NAME> passes if there is no audit by Q3")
  a market fact either party cited                   ("copper closed at $8,510")
 
CLAIMS DO NOT NEED AN ENTITY. This is the point of the step. A rule about unnamed
counterparties, a portfolio-level threshold, a process the user follows -- none of these
attach to a company, and they are exactly what the entity-keyed index above cannot hold.
Leave entity_identifiers empty in those cases rather than forcing a match.
 
SET is_user_claim CORRECTLY. True when <NAME> asserted it, false when the other party
did. Both are kept -- this records WHOSE claim it is, it does not filter. Getting it wrong
is how the other party's positions become <NAME>'s.
 
DO NOT: merge two assertions into one claim, restate the same assertion twice, invent
anything not said, or emit pleasantries and scheduling ("talk Friday", "nice to meet you").
 
TWO FIELDS ON EVERY CLAIM, both needed by retrieval:
  topic_id    -- the taxonomy id (same ids as the TOPIC tags above) the claim falls under,
                the most specific that fits. A question about "the higher-for-longer
                thesis" names no company; it reaches claims ONLY through this field.
  action_kind -- set when the claim records something <NAME> DID to a position or a
                view; null for observations and cited facts. Exactly one of:
                  INITIATED  ADDED  TRIMMED  EXITED  UPGRADED  DOWNGRADED
                  TARGET_RAISED  TARGET_CUT  STOP_TIGHTENED  STOP_LOOSENED
                  HEDGED  DEFERRED (decided to wait on a planned action)  HELD
                "<NAME> trimmed Orsted by 50bps" is TRIMMED; "<NAME> moved COIN to
                Marketweight from Underweight" is UPGRADED; "<NAME> will trim if spreads
                fall below 35%" is a rule, not an action -- null. Whether <NAME> warmed or
                cooled on a name is later read from the SEQUENCE of these, so an action
                left untagged is a step missing from that path.
 
========================================================
STEP 4 -- USER SELF-ATTRIBUTES  (only when the user reveals one)
========================================================
Separately from entities, capture stable facts the user reveals ABOUT THEMSELVES as
an investor. For each, emit a USER_ATTRIBUTE node and a HAS_ATTRIBUTE edge from USER.
 
NODE: node_type=USER_ATTRIBUTE, identifier=<one KIND slug below>, label=<the kind>.
EDGE: USER -[HAS_ATTRIBUTE]-> USER_ATTRIBUTE   (no sentiment, no topic tags).
 
KINDS -- identifier MUST be exactly one of these; never invent, never free-form.
If a self-attribute fits none of them, SKIP it -- do not force a bucket:
  RISK_TOLERANCE  -- the OVERALL amount of volatility / drawdown / loss the user is willing to
                    accept as an investor. This is about HOW MUCH risk they can stomach -- not
                    which specific risks they watch (that is RISK_FOCUS), and not their style
                    (that is PREFERENCE). Emit it WHENEVER the user signals where they sit on the
                    safety<->growth spectrum; do NOT skip a clear signal just because a PREFERENCE
                    or RISK_FOCUS is also present. ALSO set attribute_value to the closest level:
                      CONSERVATIVE -- prioritises capital preservation, safety, income, stability;
                                     wants to avoid large drawdowns; shuns speculative names.
                                     Cues: "yield plus safety", "I want stability", "I can't
                                     stomach big losses", "I stick to boring compounders".
                      MODERATE     -- accepts some volatility for growth; balanced; a mix of
                                     defensive and growth exposure.
                      AGGRESSIVE   -- chases high growth and will endure large drawdowns / high
                                     volatility / concentration for it. Cues: "I hold through any
                                     drawdown", "comfortable with big swings", "high-conviction
                                     concentrated bets", "all-in on the thesis".
                    Pick the closest; omit the level ONLY if the direction is genuinely unclear.
  RISK_FOCUS      -- the specific risks the user actively watches (concentration,
                    drawdown, inflation, rates/duration, geopolitical, liquidity, currency).
  GOAL            -- an investment objective (retire by 50, income, capital growth).
  CONSTRAINT      -- a hard limit, INCLUDING liquidity needs (no lockups > 2y, must keep
                    cash accessible, no leverage, ESG-only).
  TIME_HORIZON    -- investment horizon (short-term / multi-year / retirement).
  EXPERIENCE      -- self-described expertise (beginner ... professional).
  PREFERENCE      -- a durable style/preference (value investing, dividends, index-only).
 
The three risk-shaped kinds are DIFFERENT axes and often co-occur -- emit each that applies:
  RISK_TOLERANCE = how MUCH risk overall (a level) | RISK_FOCUS = WHICH risks they watch |
  PREFERENCE = their STYLE. "Dividends for safety" is BOTH a dividend PREFERENCE and a
  CONSERVATIVE RISK_TOLERANCE -- do not collapse them into one.
 
WORLD vs SELF -- the deciding test:
  ENTITY memory  = a third-person stance on a NAMED thing ("NVDA is overvalued").
  USER_ATTRIBUTE = a first-person self-description that holds regardless of any specific
                   instrument ("I'm risk-averse", "I want to retire by 50", "I can't lock
                   up cash for 2 years", "I'm a beginner").
 
RULES:
  1. Classification ONLY -- flag WHICH kinds are present, plus the RISK_TOLERANCE
     controlled level (attribute_value). Do NOT write any prose value/description;
     the value sentence is produced later in synthesis.
  2. Emit BOTH when warranted -- an attribute does NOT replace an entity view, and one
     self-description can reveal SEVERAL attributes at once. Emit every kind that applies:
       "I'm bullish on NVDA because it fits my aggressive risk appetite"
         -> NVDA INTERESTED_IN (BULLISH)  AND  RISK_TOLERANCE (AGGRESSIVE).
       "My whole approach is yield plus safety -- dividends are the anchor and I hold boring
        compounders through any cycle"
         -> RISK_TOLERANCE (CONSERVATIVE)  AND  PREFERENCE (dividend/income style).
       "I'm building one concentrated AI-supply-chain bet and I'll ride out any drawdown"
         -> RISK_TOLERANCE (AGGRESSIVE)  AND  RISK_FOCUS (concentration).
  3. Boundary trap -- first-person "I am / I only / I always" self-descriptions are
     ATTRIBUTES even when they name a concept:
       "I'm a value investor"  -> USER_ATTRIBUTE PREFERENCE
       "I only buy ESG names"  -> USER_ATTRIBUTE CONSTRAINT
       "I love dividends"      -> USER_ATTRIBUTE PREFERENCE
     Evaluating a concept as an external thing ("is ESG overrated?") -> ENTITY, not attribute.
  4. Must be the USER's OWN attribute -- never assistant advice. Do not invent; skip if
     merely implied. USER_ATTRIBUTE nodes are NOT tagged with topics.
  5. At most one node per kind.
 
========================================================
OUTPUT
========================================================
Return JSON only -- no commentary. Match the ExtractionResult schema exactly:
  is_trivial:            bool -- true ONLY for pure filler (greetings, one-word replies).
  extraction_confidence: float [0, 1].
  entities:              list of {node_type, identifier, label, entity_subtype?, attribute_value?, confidence?}
                         (ENTITY, TOPIC_*, and any USER_ATTRIBUTE nodes; attribute_value only on RISK_TOLERANCE)
  claims:                list of {text, exchange_id, is_user_claim, entity_identifiers?, topic_id, action_kind?}
  relations:             list of {subject_identifier, relation_type, object_identifier, sentiment?, anchors?}
                         (USER->ENTITY, ENTITY->ENTITY, ENTITY->TOPIC, and any USER->USER_ATTRIBUTE;
                          anchors only on USER->ENTITY, and only when a revisable quantity was stated)
\end{lstlisting}
 
\subsection{Synthesis}
\label{app:prompt-synth}
One call per session after extraction (\textsc{Synthesise} in Algorithm~\ref{alg:write}).
It receives the dated turns of both participants, one row per touched
(entity, topic) pair with its stored text and sentiment, and one row per
detected attribute with its stored value, and returns the new text of each.
The attribute block is omitted when no attribute was detected.
 
\Needspace{8\baselineskip}
\begin{lstlisting}[style=prompt,title={Synthesis prompt}]
You are updating a long-term financial memory system for a user after a new conversation.
 
USER: <NAME>   (graph id: <USER_ID>)
The conversation has two participants. Lines prefixed "<NAME>:" are the user whose memory
this is. Lines prefixed "(other party)" are someone else -- never the memory's subject.
 
NEW CONVERSATION:
<DATED_CONVERSATION_TURNS>
 
========================================================
ENTITY-TOPIC ROWS TO UPDATE
========================================================
<ENTITY_TOPIC_ROWS_WITH_PRIOR_STATES>
 
========================================================
USER SELF-ATTRIBUTES TO UPDATE
========================================================
<ATTRIBUTE_ROWS_WITH_PRIOR_VALUES>
 
For EACH attribute kind above, write one updated value sentence describing what
<NAME> revealed about THEMSELVES as an investor (not about any external entity):
  - FIDELITY FIRST -- anchor on what <NAME> ACTUALLY said. Preserve the concrete specifics
    verbatim where possible: named limits ("no leverage", "ESG-only"), figures and targets
    ("retire by 55", "10-15 year horizon"), the specific risks named ("concentration",
    "inflation"), the stated experience level ("20 years", "beginner"). These specifics ARE the
    value -- do not paraphrase them away into generic investor-speak.
  - DO NOT INVENT -- never introduce a limit, goal, figure, risk, or preference <NAME> did not
    state. E.g. do not add "liquidity" unless they raised it, and do not infer a goal from which
    stocks they hold. Stay minimal and close to their words.
  - EVERY kind listed above was already DETECTED as present, so it IS supported -- by this
    conversation, the prior value, or both. Locate that support and render it faithfully. NEVER
    reply that the attribute is undisclosed / unknown / not specified. If this conversation truly
    adds nothing new for a kind, restate the prior value unchanged (and if there is no prior value,
    the kind would not be listed -- so find the statement in the conversation).
  - State the value plus its "why" when they gave one. For RISK_TOLERANCE the controlled level is
    given above -- the sentence must read consistently with it.
  - Merge with the prior value -- your output fully replaces it. EVOLUTION: if it shifted vs the
    prior value, capture the change, not just the endpoint.
  - Worked contrast for a CONSTRAINT stated as "no leverage, and I only buy ESG names":
      BAD:  "<NAME> prioritizes liquidity and high-quality assets."   (invented + generic)
      GOOD: "<NAME> will not use leverage and invests only in ESG-screened companies."
  - Return one SynthesisResult entry per kind with entity_id = the KIND slug (e.g.
    "RISK_TOLERANCE"), and sentiment = null (attributes never carry sentiment).
  - Max 600 characters; use "<NAME>", never pronouns. Keep figures <NAME>
    stated exact (targets, %, limits, dates) -- the raw conversation is not stored, so a
    dropped figure is lost permanently.
 
========================================================
ONE ROW PER (ENTITY, TOPIC) -- SCOPE EACH ROW
========================================================
Each row above is a SEPARATE memory covering ONE topic for that entity. The row header
names the topic and the specific angles discussed under it.
 
  1. STAY IN SCOPE. Write only what this conversation says about THAT topic. Material
     belonging to another row is that row's job -- never repeat it.
       Row NVDA @ l2_valuation      -> the multiple, the price, what it is worth.
       Row NVDA @ l2_income_statement -> revenue, margins, segment growth.
     Writing "bullish on datacenter growth but wary at 40x" into BOTH rows is the
     failure mode this design exists to prevent: it produces two near-identical
     memories that crowd each other out of retrieval.
  2. OMIT A ROW in either of these cases -- omission is a normal outcome, not an error:
     (a) it HAS a prior memory and this conversation adds nothing new about that topic;
         the stored value then stays untouched, which is preferred over restating it or
         padding it with another topic's material.
     (b) another row for the SAME entity already carries this view. The topic rows are
         proposed mechanically from tags, so a single thesis sometimes gets split across
         two of them. When that happens, write it ONCE in the row it fits best and omit
         the other -- do NOT paraphrase the same view into both.
           e.g. "TSM has pricing power as the indispensable foundry" is ONE view: put it
           in the row whose angles match it, and drop the other row.
  3. NEVER drop an entity entirely. Every entity in the list must end up with at least
     one row. If all of an entity's rows show "prior memory: none", at least one MUST be
     written -- there is nothing stored to fall back on, so omitting them all destroys the
     memory. Where the conversation is thin, write the little it supports -- even just the
     reason the entity came up.
  4. Each row's sentence must stand alone. A reader seeing only that row must
     understand the view, so do not write "as noted above" or refer to other rows.
  5. Sentiment is per row. <NAME> can be bearish on valuation and bullish on growth
     at the same time; that is the point of separating them. Set each row's sentiment
     from ITS OWN content, not from an overall stance on the entity.
  6. Echo topic_id back exactly as given. A row whose topic_id was not requested is
     discarded.
 
========================================================
TASK
========================================================
For EACH row above, write one updated context sentence that merges prior memory
with new signal from this conversation. This is the single authoritative memory
record for this user-entity-topic triple -- your output fully replaces the prior context,
it is not appended.
 
A useful sentence captures, whenever the conversation supports it:
  - STANCE   -- what <NAME> believes about the entity, and how strongly (conviction).
  - DRIVER   -- the reasoning behind that stance: the specific angle engaged
               (valuation, growth, margins, a catalyst, management, a macro factor).
               Always pair a view with its "why".
               The row's topic IS the angle -- go deep on it rather than
               surveying every angle of the entity.
  - ANCHORS  -- concrete specifics cited: price levels, targets, multiples, dates,
               catalysts, % moves. Prefer numbers over adjectives.
  - EVOLUTION-- if the view shifted vs prior memory, capture the change, not just the
               endpoint: "was bullish pre-earnings; trimmed post-print, remains net long."
 
RULES:
  1. Output one entry per ROW you have new signal for -- exact entity_id AND topic_id
     from the list above. Omit rows with nothing new (see scoping rules).
  2. Sentences must stand alone: use "<NAME>" and entity identifiers, never pronouns.
  3. Record <NAME>'s OWN view and reasoning -- not generic facts about the entity,
     and not the other party's opinion. Their lines are shown ONLY so you can resolve
     what <NAME> is referring to. An assertion <NAME> did not take up is NOT a
     memory.
 
     EVERY FIGURE BELONGS TO WHOEVER SAID IT. Before writing any number, trigger level,
     position size or target into the sentence, locate the turn it came from:
       said in a <NAME> turn          -> it is theirs; state it plainly.
       said by the other party, and <NAME> explicitly adopted it ("I agree",
         "let's use that", "that's my level too")
                                          -> it is theirs; state it plainly.
       said by the other party, and <NAME> merely replied, discussed, or reacted
                                          -> NOT theirs. Either omit it, or attribute it
                                            ("the counterparty's $8,700 trigger").
     Engagement with an ENTITY does not transfer the other party's numbers about it.
     Worked example -- the other party says "copper is ~$10 shy of my $8,700 trigger to
     add 15% to my copper position" and <NAME> replies about copper generally:
       BAD:  "<NAME> ... is considering a 15% position increase at the $8,700 trigger."
       GOOD: "<NAME> remains bullish on copper" (their own view; the $8,700 and the
              15% are the counterparty's and belong to no memory of <NAME>).
     This is the single costliest error available here: it manufactures commitments
     <NAME> never made, and reads back as fact with no way to tell it is wrong.
 
     A sentence with no stance AND no "why" is a failure:
       BAD:  "<NAME> discussed NVDA's earnings."      (no view, no reasoning)
       BAD:  "NVDA is a leading AI chipmaker."           (generic fact, not the user's view)
       GOOD: "<NAME> is bullish on NVDA into earnings, expecting a datacenter beat to
              push past a 600 price target, but flags China export controls as the key risk."
  4. Reinforcement: if the new conversation confirms prior memory, reflect increased conviction.
  5. First time (prior memory none): summarise only what this conversation reveals -- do not invent.
  6. INTERESTED_IN entities: set sentiment to the net view, and the context sentence MUST
     read consistently with that label -- the two are reviewed side by side as a sentiment
     drift timeline, so they can never contradict.
       BULLISH -- context leans net positive (upside/long/constructive).
       BEARISH -- context leans net negative (downside/short/cautious-to-negative).
       MIXED   -- context cites MATERIAL offsetting positive AND negative drivers with no
                 clear net winner within THIS topic (e.g. under l2_valuation:
                 "cheap on FCF yield but rich on forward P/E"). Use MIXED here, NOT NEUTRAL.
       NEUTRAL -- <NAME> is engaged but genuinely expresses no directional lean (rare).
                 The context must NOT read directionally bullish/bearish if you pick NEUTRAL.
     Never emit a directionally-worded context with a NEUTRAL or opposite-direction label.
  6b. ATTRIBUTION -- WHOSE VIEW IS IT. Facts and framing may come from the assistant; the
     STANCE must be <NAME>'s. A row's sentence states what <NAME> thinks or did -- never
     what the assistant concluded, however well argued.
     The failure this exists to stop: <NAME> ASKS a question ("is AAPL good right now?"),
     the assistant answers with analysis, and the row is written as though that analysis
     were <NAME>'s own position ("...viewing the company as high quality but at a full
     valuation"). <NAME> said none of it. They asked.
     When the turn is a QUESTION, the honest sentence records the asking and the subject
     -- "<NAME> asked whether AAPL is attractive at the current price" -- not the answer.
     Whatever the assistant said is already preserved as a CLAIM; it does not need to be
     laundered into a memory to survive.
  7. MENTIONED entities (no thesis): sentiment must be null -- capture WHY the entity came up
     and how it connects to <NAME>'s other interests, rather than a stance. A null-sentiment
     context must NOT assert a bullish/bearish stance.
  8. Consider how entities relate to each other within this conversation when writing context.
  9. Max 600 characters. Be information-dense -- every clause should add signal, not
     filler. NOTHING outside this record survives: the raw conversation is not stored, so a
     figure or commitment you leave out is lost permanently, not merely summarised away.
     When the budget binds, preserve in THIS order and drop from the bottom:
       (a) NUMBERS and hard commitments stated by <NAME> -- price levels, targets,
           multiples, %, position sizes, thresholds, dates, deadlines. Keep them exact
           ("adds below 850", "trimmed at 55x", "caps NVDA at 7%"), never rounded or
           softened into words like "high" or "a lot", and never converted to another
           scale -- 8% stays "8%", not "0.08".
       (b) the stance and its direction.
       (c) the reasoning / driver behind it.
       (d) narrative colour and framing.
 
     THREE RULES FOR FIGURES, in order of precedence:
       i.   CARRY FORWARD. A figure in the prior memory that is still current must survive
            even when this conversation never mentions it. The conversation is not stored;
            dropping the figure deletes it from the system permanently. If prior memory
            says "adds below 850" and today's turns are about margins, the new sentence
            still says "adds below 850".
       ii.  REPLACE ON REVISION, and show the move. When <NAME> supersedes a figure,
            state the change rather than only the endpoint: "cut the cap from 10% to 7%".
            Never leave the old value asserted as if it were current.
       iii. STAY IN YOUR ROW. State a figure in the row whose topic it belongs to, and do
            NOT restate another row's figure. Two copies of one number drift apart the
            moment one row is updated and the other is not, and the stale copy then reads
            as a current commitment. An entry level belongs in the positioning row, a
            multiple in the valuation row -- not in both.
 
Return JSON only matching the SynthesisResult schema:
  entities: list of { entity_id, topic_id, context, sentiment }
\end{lstlisting}
 
\subsection{Quantity resolver}
\label{app:prompt-resolver}
Called only for new commitments that have stored candidates with the same
entity, kind, unit, currency and period (\S\ref{sec:construction}).
\texttt{\{user\}} is the graph owner and \texttt{\{blocks\}} holds one block
per contested figure, built by the code below; candidate indices are local
to each block. A match reuses the stored slot; \texttt{null}, a missing
decision or an out-of-range index gives the figure a new slot.
 
\Needspace{8\baselineskip}
\begin{lstlisting}[style=prompt,title={Resolver prompt}]
A memory system tracks the quantities {user} has committed to -- position sizes, caps,
stops, add/trim levels, targets, yields, estimates. A later value REPLACES an earlier one
only when both are the SAME quantity. Decide that for each new figure below.
 
SAME quantity -- the new figure revises or restates the stored one:
  - the same decision variable, about the same position, account and scenario
  - "my stop is 48" then "moved the stop to 45"; "PT 410" then "raising my target to 500"
DIFFERENT quantity -- even with the same entity, kind and unit:
  - another account, mandate, client or portfolio sleeve
  - another scenario or case (base vs downside vs upside)
  - planned vs actual (a starter size being considered vs the weight actually held)
  - a part vs the whole (the amount trimmed or added vs the resulting position)
  - a different trigger (a stop-loss vs a profit-taking trim; an add on a dip vs an add
    on a breakout), a different instrument or leg, a different horizon
  - somebody else's number
When the descriptions do not settle it, answer DIFFERENT: a wrongly merged quantity
destroys the old value's identity, a wrongly split one only costs a duplicate line.
 
{blocks}
 
Return one decision per NEW figure.
\end{lstlisting}
\Needspace{8\baselineskip}
\begin{lstlisting}[style=prompt,title={Construction of \texttt{\{blocks\}} (Python)}]
    blocks = []
    for n, (_, anchor, cands) in enumerate(contested):
        cur = f" {anchor.currency}" if anchor.currency else ""
        lines = [f"NEW {n}: {cands[0].entity_id} {anchor.kind.value} -- "
                 f"{anchor.basis or anchor.qualifier or '(no description)'} -- "
                 f"{anchor.comparator.value} {anchor.value:g} {anchor.unit.value}{cur}",
                 "  STORED quantities of this kind:"]
        lines += [f"    {j}: {c.describe()}" for j, c in enumerate(cands)]
        blocks.append("\n".join(lines))
\end{lstlisting}
 
\subsection{Router}
\label{app:prompt-router}
The only generative call at retrieval; the question embedding and, for
unresolved names, entity embeddings are separate requests. The router
returns a grounding label, not weights: the code maps it to the $\lambda$ of
Table~\ref{tab:weights}. If the output cannot be parsed, retrieval proceeds
with an empty intent whose grounding is semantic-only.
 
\Needspace{8\baselineskip}
\begin{lstlisting}[style=prompt,title={Router prompt}]
You are a financial query router for a personal memory system. Classify what this query needs so the retriever can fetch the right kinds of stored memory.
 
QUERY: <QUESTION>
 
WHO IS ASKING: <NAME>  (memory owner id: <USER_ID>)
Every memory in this store already belongs to <NAME>, so <NAME> is NEVER an entity_ids
value -- naming them there matches everything and therefore grounds nothing. The same goes
for first-person references ("my", "I") and for pronouns standing in for <NAME>.
Other people go in person_ids, NOT entity_ids: a counterparty, a colleague, an analyst,
whoever <NAME> was talking to. entity_ids is for SUBJECTS <NAME> holds views about --
companies, tickers, sectors, instruments, macro factors. The split matters because an
unresolved entity_ids value is read as "this store has nothing on that", and a person's
name never resolves to a subject even when the store answers the question perfectly.
A person named as the SUBJECT of an opinion ("what does she think of Jensen Huang?")
belongs in entity_ids; a person named as a conversation participant does not.
 
Topic taxonomy (closed vocabulary -- use exact ids only).
Prefer the most specific level (L3 > L2 > L1). Empty list if no clear topic.
l1_fundamental
  l2_income_statement: l3_revenue_growth, l3_gross_margin, l3_ebitda, l3_net_income, l3_operating_leverage, l3_pricing_power, l3_cost_structure, l3_revenue_mix
  l2_balance_sheet: l3_debt_equity, l3_cash_position, l3_working_capital, l3_goodwill, l3_covenant_risk, l3_asset_quality, l3_net_debt
  l2_cash_flow: l3_free_cash_flow, l3_capex_cycle, l3_cash_conversion, l3_dividend_coverage, l3_cash_burn, l3_operating_cash_flow
  l2_valuation: l3_pe_ratio, l3_ev_ebitda, l3_ev_sales, l3_dcf, l3_sum_of_parts, l3_comps, l3_precedent_tx, l3_nav, l3_price_target, l3_lbo, l3_fcf_yield
  l2_management_quality: l3_guidance_accuracy, l3_capital_allocation, l3_insider_ownership, l3_esg_governance, l3_compensation, l3_track_record, l3_board_composition, l3_related_party
l1_market_dynamics
  l2_price_action: l3_breakout, l3_support_resistance, l3_volume_anomaly, l3_momentum, l3_relative_strength, l3_technical_indicators
  l2_institutional_flow: l3_13f, l3_block_trades, l3_etf_flows, l3_options_flow, l3_short_interest, l3_passive_rebalancing
  l2_sentiment: l3_upgrade, l3_downgrade, l3_short_squeeze, l3_retail_sentiment, l3_market_sentiment, l3_options_sentiment, l3_consensus, l3_earnings_revision, l3_pt_revision
  l2_liquidity: l3_bid_ask, l3_market_depth, l3_adv, l3_float, l3_execution_cost
l1_corporate_actions
  l2_earnings_events: l3_earnings_date, l3_eps_surprise, l3_revenue_surprise, l3_guidance_raise, l3_guidance_cut, l3_analyst_day, l3_preannouncement, l3_earnings_call
  l2_ma_activity: l3_acquisition_target, l3_deal_synergies, l3_antitrust, l3_spinoff, l3_asset_sale, l3_merger_arb, l3_deal_premium, l3_strategic_review
  l2_capital_structure: l3_buyback, l3_convertible, l3_dividend_init, l3_rights_issue, l3_secondary_offering, l3_debt_refinancing, l3_credit_rating
  l2_management_changes: l3_ceo_change, l3_cfo_hire, l3_board_change, l3_activist, l3_founder_departure
l1_quant_risk
  l2_factor_exposure: l3_quality_factor, l3_momentum_factor, l3_low_vol, l3_value_factor, l3_growth_factor, l3_size_factor, l3_profitability
  l2_portfolio_risk: l3_beta, l3_var, l3_correlation, l3_drawdown, l3_concentration, l3_tail_risk, l3_tracking_error, l3_stress_test, l3_sharpe
  l2_earnings_quality: l3_accrual_ratio, l3_rev_recognition, l3_audit_quality, l3_restatement, l3_non_gaap, l3_financial_distress
l1_macro_sector
  l2_macro_environment: l3_interest_rates, l3_fed_policy, l3_yield_curve, l3_inflation, l3_employment, l3_gdp, l3_recession, l3_credit_spreads
  l2_sector_dynamics: l3_market_share, l3_sector_pricing, l3_tam, l3_moat, l3_supply_chain, l3_inventory_cycle, l3_regulation, l3_disruption, l3_consolidation
  l2_geopolitical_risk: l3_tariffs, l3_sanctions, l3_regulatory_probe, l3_trade_war, l3_export_controls, l3_country_risk, l3_elections
  l2_commodities_fx: l3_energy_prices, l3_metals_prices, l3_agricultural_commodities, l3_fx_rates, l3_commodity_cycle, l3_fx_hedging
 
Return JSON matching the QueryIntent schema. Be decisive about query_type, and set needs_history only when the past state is genuinely required -- a present-tense question answered from a superseded state is a wrong answer, not a partial one.
\end{lstlisting}
 
\subsection{Answer}
\label{app:prompt-answer}
Shared by all methods; only \texttt{<CONTEXT>} differs (the memory context,
the full transcript, or the gold evidence turns). No reference answer is
included. A reply of exactly ``No information available in memory.'' is
recorded as a decline. Gemini receives the output schema through its native
structured-output interface; Nemotron receives a strict schema and a short
appended instruction to return only the JSON object.
 
\Needspace{8\baselineskip}
\begin{lstlisting}[style=prompt,title={Answer prompt}]
You are answering a question from a personal financial memory system.
 
CONTEXT below is everything the system remembers about <NAME>. It is not a document and
not a transcript -- it is a lossy memory record, and it is deliberately incomplete. Plenty
of true things about <NAME> are simply not in it.
 
QUESTION: <QUESTION>
 
CONTEXT:
<CONTEXT>
 
HOW TO ANSWER
 
1. SUPPORT CHECK. For each part of your answer, locate the specific line in CONTEXT that
   supports it. If you cannot point to one, you do not have the answer -- say so. Do not
   fill a gap by reasoning from what a person like <NAME> would plausibly do.
 
2. FALSE PREMISE. A question can presuppose something that never happened -- "what did
   <NAME> ultimately execute", "what leverage did <NAME> use", "what did <NAME> do with
   the proceeds". If CONTEXT does not show that it happened, the question is unanswerable.
   Do NOT answer with the nearest similar thing <NAME> did do; that is a confident wrong
   answer, which is worse than no answer.
 
3. WRONG SUBJECT -- and read this one carefully, because it cuts finely.
   Unanswerable: what ANOTHER PERSON thinks, holds, decided, or did. CONTEXT is <NAME>'s
     memory and holds no record of anyone else's positions or intentions. "What is her
     colleague's target", "what did he do with the proceeds" -- decline.
   ANSWERABLE: a figure that originated elsewhere but which <NAME> CITED, relayed, or
     reacted to, when CONTEXT records it. A bank's estimate, a conference statistic, a
     company's guidance, a market print -- these are legitimately in <NAME>'s memory and
     EXACT FIGURES marks each with who stated it. "What EBITDA cut did the broker apply",
     "what default rate was flagged" -- answer these from CONTEXT, and say whose figure it
     is. Do NOT decline merely because a third party is named in the question.
   The test is whose STATE OF MIND is being asked about, not whose mouth a number came
   out of.
 
4. WHICH VALUE -- decide this before reading off any number. Most quantities here were
   revised, so CONTEXT usually holds several values for the thing being asked about.
     - The question pins a MOMENT: a date, "initially", "at first", or an event or
       condition ("after the Fed signal", "when he trimmed", "following the guide-down",
       "on the February sale"). Answer with the value tied to THAT moment -- the dated
       statement or series entry that matches the event -- even though a later value
       exists. The latest value is the wrong answer to a question about an earlier one.
     - The question asks about NOW ("currently", "latest", "where does it stand") or
       about how something CHANGED: use the latest value, or walk the series in order.
     - The question pins nothing: prefer the dated statement whose wording matches the
       question's own details (the same action, size, counterpart figure) over a value
       that merely shares the entity.
   For the same quantity at the same date, EXACT FIGURES beat the prose, which may be
   rounded or loosely restated. They do NOT beat a dated statement about a different
   moment -- a different date is a different fact, not a less exact one.
 
5. PREVIOUSLY HELD states are what <NAME> used to think. They are the right source when
   rule 4 points at the past; never present one as <NAME>'s current view.
 
6. HOW MANY DAYS. If the QUESTION itself gives both dates, the answer is the difference
   of those two dates; CONTEXT only has to confirm the events happened. Otherwise, when
   dated lines carry a "day N" marker, the days between two events
   is the difference of their markers -- find the line for EACH event the question names,
   then subtract. Do not subtract calendar dates by hand, and do not use a line merely
   because it is dated: it must be the event the question describes. A statement made in
   a conversation about something that happened "yesterday" or "last week" happened
   that much before the line's date.
 
7. WARMED OR COOLED means CONVICTION in the thesis, not position size. Read it from the
   whole record in date order -- statements of view, ratings, targets, what <NAME> kept
   returning to -- and then from the actions. Strongest signals: an upgrade or downgrade,
   a target raised or cut, a thesis restated with more or less confidence, a name called
   a top pick or dropped from consideration. Sizing moves are weaker and ambiguous: a
   trim can be risk management or profit-taking inside a thesis that is strengthening,
   and an add can be mechanical. Do not call a trim "cooling" when the stated view in the
   same period is unchanged or firmer. Where signals conflict, the stated view decides
   and the sizing move is mentioned as context. Where the path reversed, answer where it
   ENDED relative to where it began, and say so.
 
If rule 1, 2 or 3 applies, reply with exactly this and nothing else:
No information available in memory.
 
Otherwise fill EVERY field:
 
  value        -- the bare answer, nothing around it. Several values are separated by
                 "; " in the order the question asks.
                     "What is her CDS trigger?"        -> 550bps
                     "...target widen, and what multiple?" -> 15-20bps; 2-3x
                     "How many days between X and Y?"  -> 17 days
  answer       -- the same content as one or two sentences, naming EVERY element the
                 question asks for. For a question about how something EVOLVED or about
                 an OVERALL APPROACH, this must characterise it, not just cite a number.
                     -> Liquidity started at $7.8B against her $4B threshold and rose to
                        $9.4B; the $9.4B unrestricted cash moved her back to Marketweight.
  abstained    -- false here; true only in the decline case above.
  supported_by -- the CONTEXT lines you relied on, quoted short.
 
Reuse CONTEXT's figures verbatim in both. Do not pad `value`, and do not truncate
`answer` to a bare figure -- they are different renderings on purpose.
 
Answer the whole question. If it has two parts and CONTEXT supports only one, give that
part and note the other is not in memory; abstained stays false -- do not decline the
entire question over a missing half.
\end{lstlisting}

\subsection{Judge}
\label{app:prompt-judge}
Used for answerable questions. \texttt{\{candidate\}} is the reader's
\texttt{answer} field (its \texttt{value} field if \texttt{answer} is empty).
The judge sees the question, the reference and the candidate, but neither
the context nor the method. Verdicts score 1 (correct), 0.5 (partial) and 0
(incorrect); an unrecognised verdict scores 0. Declines on unanswerable
questions are read from the reader's \texttt{abstained} field, without the
judge.
 
\Needspace{8\baselineskip}
\begin{lstlisting}[style=prompt,title={Judge prompt}]
You are grading one answer from a memory system against a reference answer.
 
QUESTION:  {question}
REFERENCE: {gold}
CANDIDATE: {candidate}
 
Grade ONLY whether the candidate conveys the same substantive answer as the reference.
 
  IGNORE: length, phrasing, word order, extra correct detail, and any hedging.
  A terse candidate that gives the right figures is CORRECT even if the reference is a
  full sentence. A verbose candidate is CORRECT if the right answer is in there.
 
  PARTIAL: the question asks for several things and the candidate gets some but not all.
  INCORRECT: a different value, the wrong entity, the wrong person's figure, a declined
  answer where the reference has content, or an answer that contradicts the reference.
 
Numbers must match to be correct. $9.4B and $9.4 billion are the same; $9.4B and $7.8B
are not.
\end{lstlisting}
The prompt is followed by an instruction to return only a JSON object
with the fields \texttt{verdict} (\textsc{correct}, \textsc{partial} or
\textsc{incorrect}) and \texttt{reason} (one short clause).

\end{document}